\documentclass[10pt,twocolumn,letterpaper]{article}

\usepackage[pagenumbers]{cvpr}

\usepackage{xspace}

\definecolor{cvprblue}{rgb}{0.21,0.49,0.74}
\definecolor{dgreen}{rgb}{0.0, 0.39, 0.0}
\usepackage[pagebackref,breaklinks,colorlinks,allcolors=cvprblue]{hyperref}
\usepackage{booktabs}
\usepackage{multirow}
\usepackage{siunitx}
\usepackage[normalem]{ulem}
\usepackage{caption}
\usepackage{bbm}
\usepackage{amsmath}
\usepackage{changepage}
\usepackage{enumitem}
\usepackage{pdflscape}

\newcommand{\method}{Rewis3d\xspace}

\newcommand{\myparagraph}[1]{\vspace{3pt}\noindent{\bf #1}}
\newcommand\blfootnote[1]{\begingroup\renewcommand\thefootnote{}\footnote{#1}\addtocounter{footnote}{-1}\endgroup}

\def\confName{CVPR}
\def\confYear{2026}

\title{\method: Reconstruction Improves Weakly-Supervised Semantic Segmentation}

\author{Jonas Ernst\textsuperscript{*1,2}, Wolfgang Boettcher\textsuperscript{*2}, Lukas Hoyer\textsuperscript{3}, Jan Eric Lenssen\textsuperscript{2}, Bernt Schiele\textsuperscript{2}\\
\textsuperscript{1}Saarland University \, \textsuperscript{2}Max Planck Institute for Informatics, SIC \, \textsuperscript{3}ETH Zurich
\\
\tt\small \{jernst, wolfgang.boettcher, jlenssen, schiele\}@mpi-inf.mpg.de \\
\tt\small lhoyer@vision.ee.ethz.ch}

\begin{document}

\twocolumn[{
  \maketitle
  \vspace{-2em}
  \begin{center}
    \includegraphics[width=\linewidth]{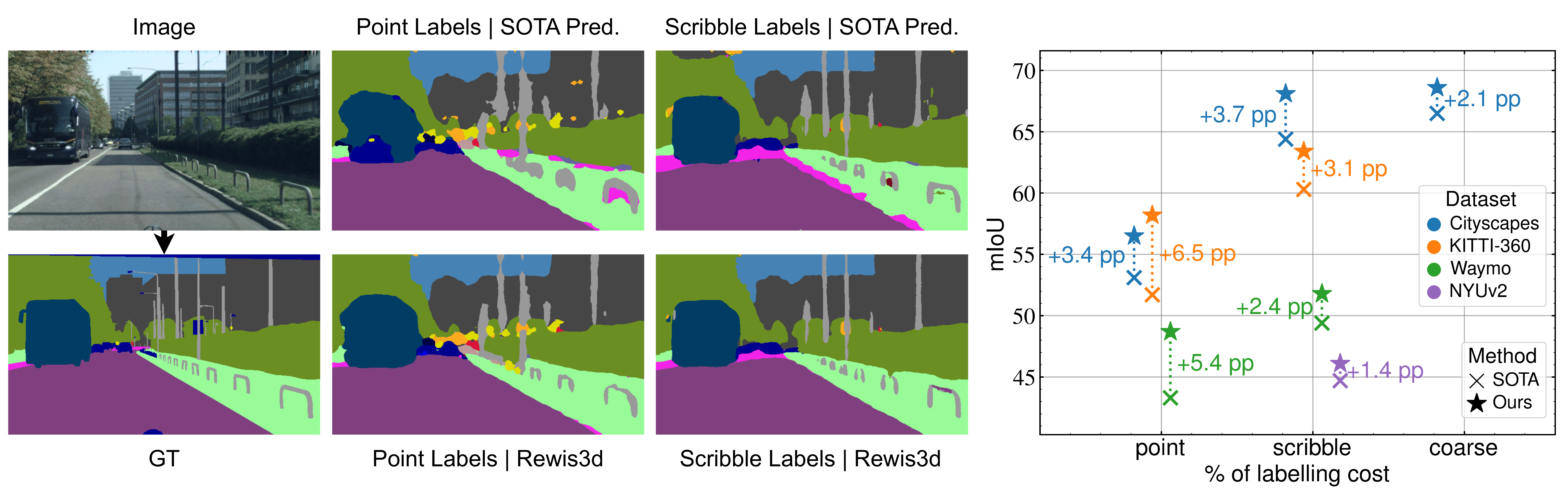}
    \\[0.5ex]
    \vspace{-0.3cm}
    \captionsetup{hypcap=false}
    \captionof{figure}{\textbf{\method} -- \emph{Left:} Our method~(Revis3d) greatly improves performance for weakly supervised segmentation, trained with point and scribble labels. Notably, we improve robustness to scale changes in objects and more precise class boundaries. \emph{Right:} We consistently outperform previous SOTA methods on a range of datasets and a variety of sparse annotations by significant margins.}
    \label{fig:teaser}
  \end{center}
}]
\begin{abstract}
We present \method, a framework that leverages recent advances in feed-forward 3D reconstruction to significantly improve weakly supervised semantic segmentation on 2D images. Obtaining dense, pixel-level annotations remains a costly bottleneck for training segmentation models. Alleviating this issue, sparse annotations offer an efficient weakly-supervised alternative. However, they still incur a performance gap. To address this, we introduce a novel approach that leverages 3D scene reconstruction as an auxiliary supervisory signal. Our key insight is that 3D geometric structure recovered from 2D videos provides strong cues that can propagate sparse annotations across entire scenes. Specifically, a dual student–teacher architecture enforces semantic consistency between 2D images and reconstructed 3D point clouds, using state-of-the-art feed-forward reconstruction to generate reliable geometric supervision. Extensive experiments demonstrate that \method achieves state-of-the-art performance in sparse supervision, outperforming existing approaches by 2-7\% without requiring additional labels or inference overhead. Our code will be released upon acceptance of the paper. \blfootnote{* Equal contribution.}\vspace{-1.5em}
\end{abstract}
\section{Introduction}
\label{sec:intro}

Semantic segmentation has achieved remarkable progress through deep learning, enabling critical applications in autonomous driving, robotics, and medical imaging~\cite{yang_semantic_2023, rizzoli_multimodal_2022, yao_cnn_2024, boettcher2023lidar}. However, these advances heavily rely on large-scale datasets with dense, pixel-accurate annotations, resources that are prohibitively expensive and time-consuming to obtain~\cite{sun_scalability_2020, cordts_cityscapes_2016, boettcher_scribbles_2024}.
A potential mitigation strategy is weakly-supervised semantic segmentation (WSSS), as it allows leveraging incomplete or imprecise annotations. Sparse spatial annotations such as points~\cite{bearman2016s}, scribbles~\cite{boettcher_scribbles_2024}, and coarse labels~\cite{cordts_cityscapes_2016} offer a compelling trade-off, providing explicit spatial localization while requiring only a fraction of the labeling effort needed for dense masks.
In this work, we present an approach to significantly improve segmentation quality for all WSSS settings (see Fig.~\ref{fig:teaser} for multiple label types and datasets) without requiring more labels or additional computation during inference.

Current approaches to bridging the weak-to-full supervision gap~\cite{10132119,Suri_2023_ICCV, bearman2016s, boettcher_scribbles_2024}, such as SASFormer~\cite{su_sasformer_2023} or TreeEnergy~\cite{liang_tree_2022}, introduce specialized architectures and loss functions to propagate information from labeled to unlabeled pixels. While effective, these techniques struggle to fully compensate for the limited supervisory signal, especially in geometrically complex outdoor scenes.
We propose a fundamentally different strategy: leveraging reconstructed 3D geometric structure as an auxiliary supervisory signal to enhance 2D weakly-supervised segmentation. Recent advances in feed-forward 3D reconstruction~\cite{wang_dust3r_2024, wang_vggt_2025, keetha2025mapanything} enable the recovery of high-fidelity 3D point clouds directly from casual 2D video sequences, without requiring specialized sensors like LiDAR. This breakthrough allows us to inject geometric constraints into the learning process while maintaining a purely 2D inference pipeline. The core principle of our method is that 3D geometry provides complementary cross-view consistency constraints that propagate sparse supervision across entire scenes: when an object is annotated with a scribble, point, or coarse label in one view, its 3D structure enables knowledge transfer to all other views in which it appears. 

\begin{figure}[t]
    \centering
    \vspace{0.2cm}
    \includegraphics[width=\linewidth]{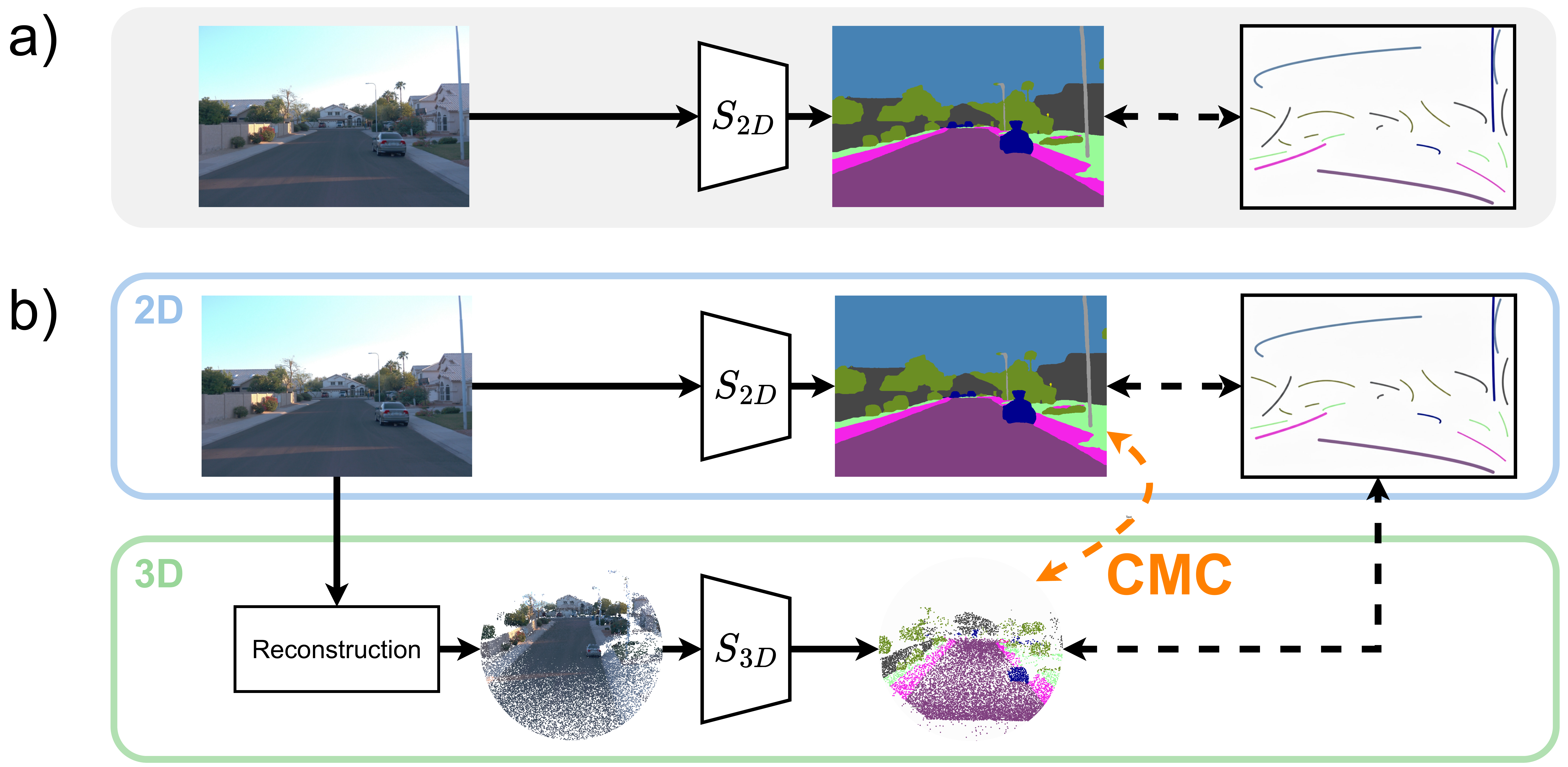}
    \vspace{-0.5cm}
    \caption{\textbf{Conceptual overview of weakly-supervised segmentation approaches.} 
    (a) Traditional methods rely solely on sparse 2D annotations, limiting supervision propagation. 
    (b) Our proposed method \method introduces a 3D branch, enforcing cross-modal consistency (CMC) between 2D predictions and 3D predictions from reconstructed geometry.}
    \vspace{-0.3cm}
    \label{fig:concepts_overview}
\end{figure}

To implement this concept, we introduce \method, a framework featuring a novel dual student–teacher architecture that enforces bidirectional consistency between 2D image-based and 3D geometry-based segmentation. Within \method, we define a cross-modal consistency\,(CMC) loss that aligns the predictions across modalities, effectively bridging the gap between reconstructed geometry and weak 2D supervision (c.f. Fig.~\ref{fig:concepts_overview}). Addressing the inevitable noise in both reconstructed geometry and weak annotations, we propose dual confidence filtering and view-aware sampling strategies that prioritize reliable 2D–3D correspondences and suppress erroneous pseudo-labels. We extend the typical evaluation scenario of WSSS methods to additional large-scale, scene-centric datasets, such as Waymo~\cite{sun_scalability_2020} and KITTI-360~\cite{liao_kitti-360_2023}, and show that our approach improves on the state of the art in mIoU by 2-7\%, achieving consistent and significant gains across a variety of datasets and supervision types. 

In summary, our main contributions are as follows:
\begin{itemize}
\item We present \method    , the first weakly-supervised framework to integrate sparse 2D annotations with 3D geometry reconstructed solely from 2D images, proving geometry as a powerful supervisory signal.
\item We introduce a novel dual student-teacher mechanism with confidence-guided filtering and view-aware sampling to ensure robust 2D–3D alignment and knowledge transfer.
\item Using only sparse supervision, our method achieves state-of-the-art results on several datasets, outperforming existing WSSS methods.
\end{itemize}

\section{Related Work}
\label{sec:related_work}
\paragraph{Weakly-Supervised Segmentation.}
Weakly supervised semantic segmentation (WSSS) aims to reduce the annotation burden of dense pixel-level labels by leveraging weaker or incomplete forms of supervision. Among these, sparse annotations, such as points~\cite{bearman2016s, tian2020weakly, zhao2020weakly}, scribbles~\cite{lin_scribblesup_2016, boettcher_scribbles_2024}, and coarse polygons~\cite{das2023urban,wang2022label}, offer a practical compromise between annotation efficiency and spatial precision. Points provide minimal but explicit localization, scribbles capture regional structure, and coarse polygons represent a denser yet still lightweight annotation form. In contrast, image-level labels~\cite{chen2025weakly,zhou2016learning, wei2017object}, which indicate only the presence of object categories without spatial cues, are less effective for scene-centric datasets where multiple object instances and overlapping categories demand precise localization to achieve meaningful segmentation.

Early sparse-label approaches, such as ScribbleSup~\cite{lin_scribblesup_2016}, used expectation–maximization to iteratively refine pseudo-labels. Later methods incorporated graph-based regularization~\cite{tang_normalized_2018, tang_regularized_2018}, self-supervised constraints~\cite{valvano_self-supervised_2021}, and uncertainty-aware refinement~\cite{xu_scribble-supervised_2021}. Recent transformer-based models like SASFormer~\cite{su_sasformer_2023} leverage self-attention to propagate sparse supervision across the image, while Tree Energy Loss (TEL)~\cite{liang_tree_2022} captures hierarchical relationships via minimum spanning trees to generate coarse-to-fine pseudo-labels.
More recently, \citet{boettcher_scribbles_2024} showed that mean teacher frameworks~\cite{tarvainen_mean_2017} remain competitive for sparse supervision, effectively propagating information from limited annotations to unlabeled regions.
Our work builds on this foundation by incorporating 3D geometric supervision to further improve consistency and segmentation quality.

\myparagraph{Weakly Supervised 3D Segmentation.}
The high cost of annotating 3D point clouds has spurred research into weakly supervised 3D segmentation~\cite{gao2025lidar}. Common strategies project 3D points onto 2D images to exploit abundant 2D labels~\cite{miao_weakly_2023, chen_weakly_2025} or adapt sparse annotations such as scribbles directly in 3D~\cite{zhang_growsp_2023, unal_scribble-supervised_2022, unal_bayesian_2025}. Recent work further explores cross-modal guidance, transferring knowledge from unlabeled 2D images via association learning~\cite{sun2024image} or aligning 3D points with textual semantics through vision-language supervision~\cite{xu20243d}. Many methods extend established 2D frameworks, such as the Mean Teacher consistency model, to the 3D domain~\cite{unal_scribble-supervised_2022, unal_bayesian_2025}. In contrast, our approach employs 3D reconstruction purely as an auxiliary supervisory signal to enhance 2D segmentation during training, while inference remains entirely in 2D.

\myparagraph{Learning-based Multi-View Stereo.}
Recent advances in multi-view reconstruction enable robust joint prediction of depth, camera parameters, and point maps in a single forward pass. MVSNet~\cite{yao_mvsnet_2018} pioneered the use of deep features for depth estimation, inspiring a new generation of methods that reconstruct dense point clouds from uncalibrated images. More recent approaches, such as DUSt3R~\cite{wang_dust3r_2024} and its successors~\cite{leroy_grounding_2025, zhang_monst3r_2025}, estimate point clouds in a canonical space, though global alignment remains necessary beyond image pairs. State-of-the-art models like VGGT~\cite{wang_vggt_2025} and MapAnything~\cite{keetha2025mapanything} generalize this concept into unified transformer-based frameworks for metric-scale 3D reconstruction, with the latter supporting additional inputs such as intrinsics and poses. These feed-forward MVS methods highlight the increasing ability of learning-based models to infer accurate 3D geometry without classical optimization or explicit 3D sensors, which we utilize in this work.

\myparagraph{Cross-Modal Consistency Learning.}
Cross-modal learning leverages complementary information from different modalities to improve task performance. In fully supervised settings, depth information has been shown to enhance 2D segmentation~\cite{hazirbas_fusenet_2017}. Under weak supervision, some methods use 3D data to refine 2D predictions~\cite{sun_3d_2020, yu_boosting_2024}, but often require 3D data at inference. Methods like 2DPASS~\cite{yan_2dpass_2022} and \citet{unal_2d_2024} distill 2D knowledge into 3D networks for LiDAR segmentation. In contrast, our proposed approach transfers geometric knowledge from reconstructed 3D to 2D during training only, enabling robust and efficient 2D-only inference without expensive 3D sensors. This bidirectional consistency between modalities, weighted by both reconstruction and prediction confidence, distinguishes our work from prior cross-modal approaches.

\section{\method}
\label{sec:methodology}
\begin{figure}
    \centering
    \vspace{0.1cm}
    \includegraphics[width=\linewidth]{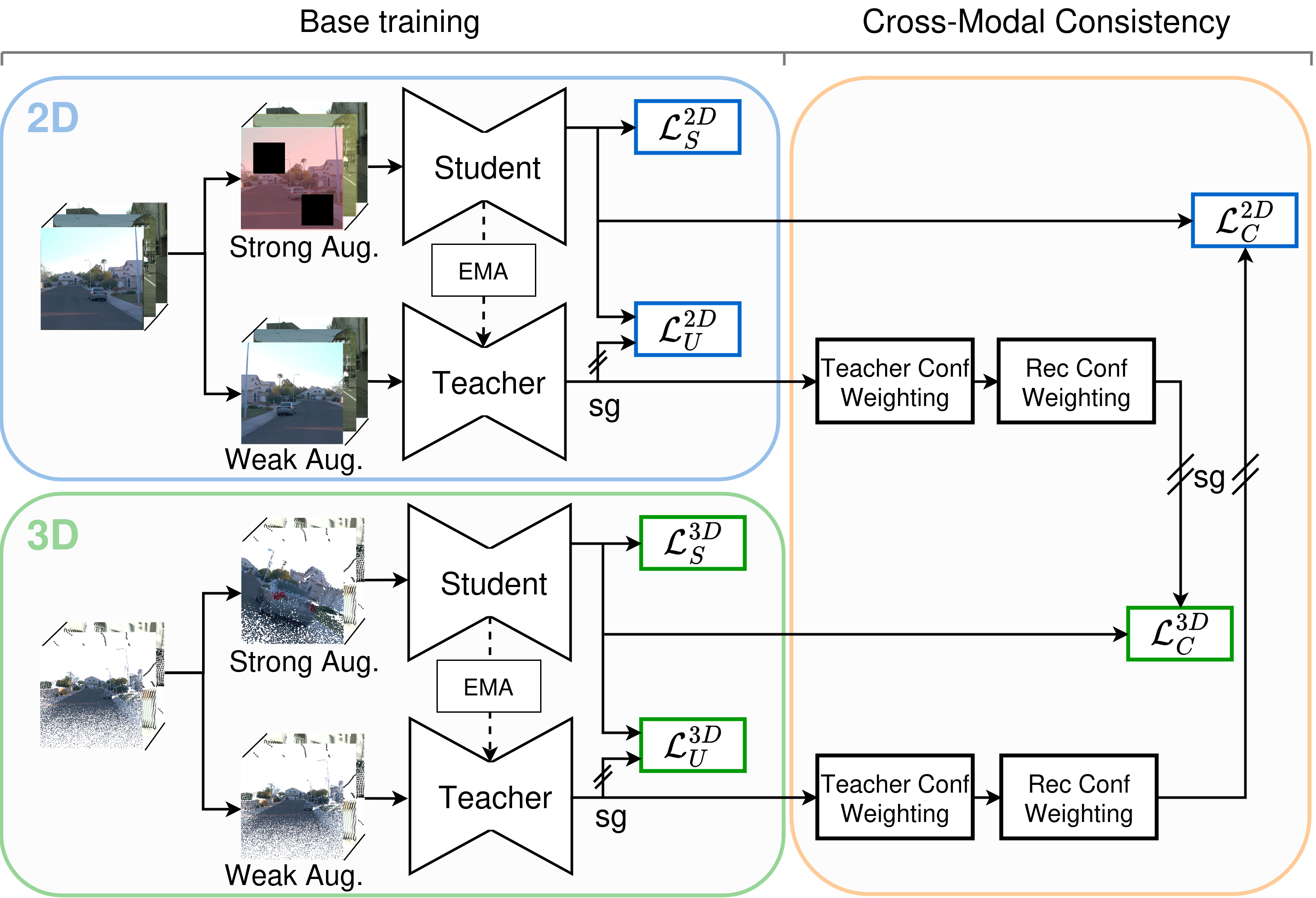}
    \vspace{-0.5cm}
    \caption{\textbf{Overview of the training pipeline.} Our framework operates in two stages. \emph{Base Training} (blue and green) establishes independent student-teacher setups for each modality using sparse supervision. \emph{Cross-Modal Consistency} (orange) introduces our core contribution: bidirectional knowledge transfer where the teacher of one modality supervises the student of the other, weighted by our dual confidence mechanism leveraging prediction certainty and reconstruction quality.}
    \vspace{-0.4cm}
    \label{fig:approach_overview}
\end{figure}

Our work improves 2D semantic segmentation from sparse annotations by leveraging 3D geometric information reconstructed from video sequences. We propose \method, illustrated in~\cref{fig:approach_overview}, that enforces bidirectional consistency between the semantic predictions for 2D images and 3D point clouds. Crucially, the 3D geometry is generated as a pre-processing step from 2D image sequences, and the final inference is performed using only a 2D image, making our approach broadly applicable without specialized 3D sensors or requiring specific segmentation architectures.

\subsection{Framework Overview}
Our framework consists of three key components that work in synergy: (1) a 2D segmentation branch, (2) a 3D segmentation branch, and (3) Cross-Modal Consistency (CMC) that enables bidirectional knowledge transfer between modalities. Each branch employs a Mean Teacher~\cite{tarvainen_mean_2017} architecture with student-teacher structures. The CMC component introduces a bidirectional consistency loss, where the teacher model from each modality generates confidence-weighted pseudo-labels to supervise the student model of the complementary modality, enabling mutual knowledge transfer between the 2D and 3D domains.

\subsection{3D Scene Reconstruction and Preprocessing}
To generate the 3D data used for supervision, we employ MapAnything~\cite{keetha2025mapanything}, a state-of-the-art multi-view stereo model that reconstructs dense, metric point clouds \mbox{$P = \{p_i\}$} and per-point reconstruction confidences $c_i^{\text{rec}}$ from 2D video sequences in a single forward pass. Unlike methods requiring post-processing optimization~\cite{wang_dust3r_2024}, \mbox{MapAnything} directly outputs camera parameters, depth maps, and point clouds, making it highly suitable for our pipeline.

\myparagraph{View-Aware Point Cloud Sampling.} Processing fully reconstructed point clouds derived from long captured sequences (often 60M+ points from 200+ images) is computationally prohibitive. Furthermore, our cross-modal consistency (CMC) loss operates on a per-image basis, requiring a dense set of 2D-3D correspondences between a target image and the points visible within its field-of-view. A simple random sampling of the entire 60M+ point scene (e.g., to 120K points) cannot satisfy this. It would yield an average of only $\sim$140 corresponding points per image, which is far too sparse to effectively train the consistency loss.

To solve this, we propose a \emph{view-aware sampling strategy} that generates a dedicated 120K-point subsample for each target image, balancing dense correspondence with global context. For a given image, its subsample is created by: (1) sampling 60\% (72K points) exclusively from the subset of points derived from that specific view, ensuring a rich set of 2D-3D correspondences for the CMC loss, and (2) sampling the remaining 40\% (48K points) from the surrounding scene within a spatial radius, providing crucial context for robust 3D segmentation. This strategy guarantees $\sim$72K correspondences for the cross-modal loss while maintaining holistic scene understanding for the 3D branch.

\begin{figure}[tb]
    \centering
    \includegraphics[width=\linewidth]{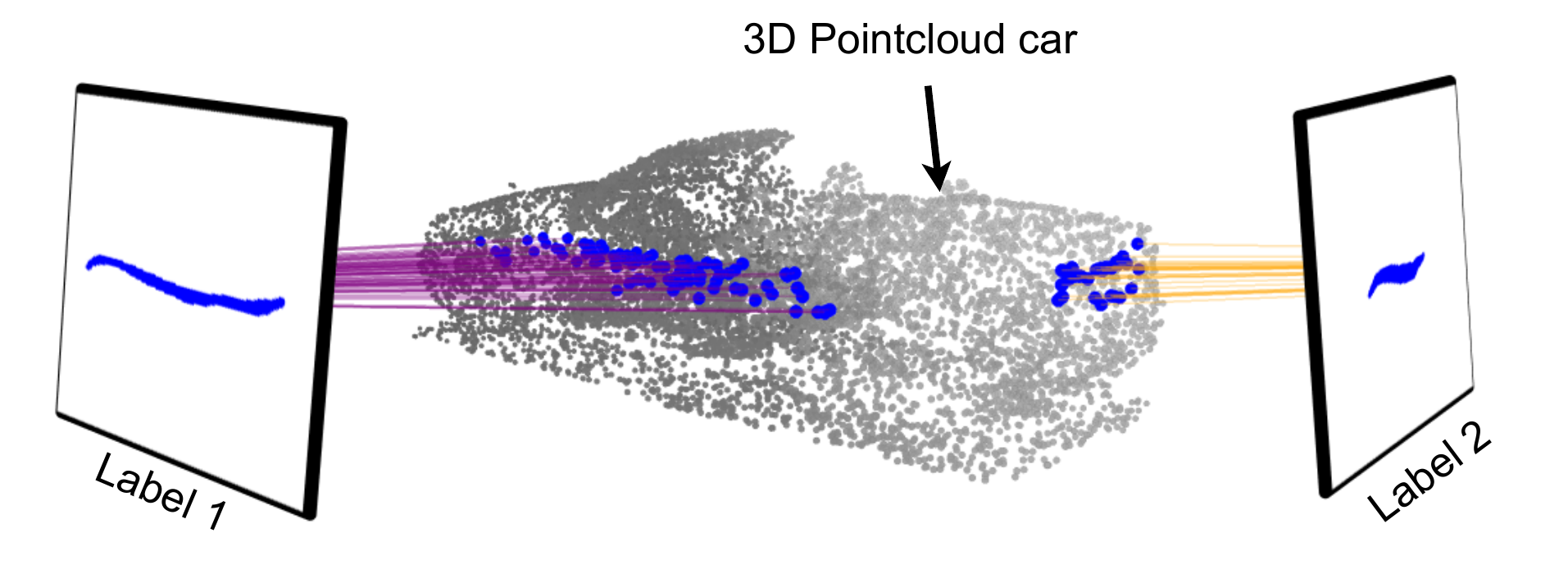}
    \vspace{-0.6cm}
    \caption{\textbf{Sparse label accumulation.} Firstly, an image sequence is unprojected to a 3D point cloud via a multi-view reconstruction model. Subsequently, we establish correspondences between the 3D points and the 2D pixels in the source images. This allows for label accumulation in the 3D space, and by projection, also in the 2D images.}
    \label{fig:projection}
    \vspace{-0.4cm}
\end{figure}

\myparagraph{3D Label Generation.} As illustrated in~\cref{fig:projection}, we generate sparse 3D labels by directly transferring annotations during the point unprojection. Since each 3D point in our cloud originates from a single 2D pixel, we assign its label in a direct one-to-one mapping. If a source 2D pixel contains an annotation, its corresponding unprojected 3D point is assigned that semantic class. Conversely, if the pixel is unlabeled, the resulting 3D point is also marked as unlabeled. By applying this process across all source images, we effectively aggregate the sparse 2D annotations from every view into a single, unified, and sparsely-labeled 3D point cloud. This sparse set of 3D labels is used for the supervised loss $\mathcal{L}_S^{3D}$, while the vast majority of unlabeled points, originating from unlabeled pixels, are supervised through teacher pseudo-labels, analogous to the 2D case.

\subsection{Dual Student-Teacher Architecture}
Our framework consists of a 2D and 3D branch, each with identical student-teacher structures. Both branches adopt a Mean Teacher~\cite{tarvainen_mean_2017} setup. This architecture is particularly effective for two reasons: (1) it enhances robustness in weakly-supervised settings by using teacher predictions as reliable pseudo-labels, and (2) the teacher's weights, updated as an Exponential Moving Average (EMA) of the student's weights, provide stable supervision for our cross-modal loss. Formally, teacher weights
\begin{equation}
    \boldsymbol{\theta}_t^{\text{teacher}} \hspace{0.1cm} \leftarrow \hspace{0.1cm} \alpha \boldsymbol{\theta}_{t-1}^{\text{teacher}} + (1 - \alpha) \boldsymbol{\theta}_t^{\text{student}}
\end{equation}
are updated at each step $t$, where we set $\alpha = 0.99$ in our experiments.

Within each branch, the student is trained with a supervised cross-entropy loss $\mathcal{L}_{S}$ on labeled regions and an unsupervised consistency loss
\begin{equation}
    \mathcal{L}_U = D_{\text{KL}}\left( \sigma(z^t_{\mathcal{U}}) \,\|\, \sigma(z^s_{\mathcal{U}}) \right)
\end{equation}
on unlabeled regions using teacher pseudo-labels, where $z^s_{\mathcal{U}}$ and $z^t_{\mathcal{U}}$ denote student and teacher logits on unlabeled pixels $\mathcal{U}$, $D_{\text{KL}}$ the Kullback–Leibler divergence, and $\sigma(\cdot)$ is the softmax function. To ensure high-quality pseudo-labels, we compute a confidence weight
\begin{equation}
    w_t = \frac{1}{|\mathcal{U}|}\sum_{i \in \mathcal{U}} \mathbf{1}\left[\max_c \sigma(z^t_i)_c \geq \tau\right]
\end{equation}
representing the fraction of pixels where the teacher's maximum class probability exceeds threshold $\tau$. The training objective for base training
\begin{equation}
    \mathcal{L} = (1 - \beta) \mathcal{L}_S + \beta \, w_t \, \mathcal{L}_U
\end{equation}
combines both supervised and unsupervised components, where $\beta$ balances the contribution of labeled and unlabeled regions, and $w_t$ adaptively scales the consistency loss based on pseudo-label quality.

\begin{table*}[htb]
    \centering
    \caption{
        \textbf{Semantic segmentation results with scribble supervision.} We report mean Intersection-over-Union (mIoU, \%) and the percentage of the supervision gap closed between scribble-supervised and fully-supervised baselines (SS/FS). Our proposed \method framework, which leverages reconstructed 3D geometry for cross-modal consistency, achieves state-of-the-art performance and consistently outperforms existing weakly supervised semantic segmentation (WSSS) methods across all datasets.
    \textbf{Bold} indicates the best scribble-supervised result, and \underline{underline} the second best.
    }
    \label{tab:main_results_2d}
    \vspace{-0.2cm}
    \resizebox{0.9\textwidth}{!}{\begin{tabular}{l l l c @{\hspace{0.9cm}} S[table-format=2.1] S[table-format=2.1] S[table-format=2.1] S[table-format=2.1] S[table-format=2.1] S[table-format=2.1]}
            \toprule
            \multirow{2}{*}{\textbf{Method}} & \multirow{2}{*}{\textbf{Ours}} & \multirow{2}{*}{\textbf{Backbone}} & \multirow{2}{*}{\textbf{3D Supervision}} & \multicolumn{2}{c}{\textbf{Waymo}} & \multicolumn{2}{c}{\textbf{KITTI-360}} & \multicolumn{2}{c}{\textbf{NYUv2}} \\
            \cmidrule(r){5-6} \cmidrule(lr){7-8} \cmidrule(lr){9-10}
            & & & & {mIoU} & {SS/FS (\%)} & {mIoU} & {\SS/FS (\%)} & {mIoU} & {SS/FS (\%)} \\
            \midrule
            Fully Supervised & & Segformer-B4 & \textendash & 59.0 & \textendash & 68.4 & \textendash & 51.1 & \textendash \\
            \midrule
            EMA & & Segformer-B4 & \textendash & 49.4 & 83.7 & 60.3 & 88.2 & 42.9 & 84.0\\
            SASFormer~\cite{su_sasformer_2023} & & Segformer-B4 & \textendash & 37.8 & 64.1 & 46.4 & 67.8 & \underline{44.7} & 87.5\\
            TEL~\cite{liang_tree_2022} & & DeepLabV3+ & \textendash & 42.4 & 71.9 & 59.2 & 86.6 & 38.3 & 75.0\\
            \midrule
            \textbf{Ours (Real 3D)} & \checkmark & Segformer-B4 & LiDAR/Depth & \underline{51.8} & \underline{87.8} & \underline{61.7} & \underline{90.2} & \underline{44.7} & \underline{87.6} \\
            \textbf{Ours (Recon)} & \checkmark & Segformer-B4 & Recon. & \textbf{53.3} & \textbf{90.3} & \textbf{63.4} & \textbf{93.4} & \textbf{46.1} & \textbf{90.2} \\
            \bottomrule
        \end{tabular}}
    \vspace{-0.3cm}
\end{table*}

\subsection{Weighted Cross-Modal Consistency}
A core component of our method is a bidirectional Cross-Modal Consistency (CMC) loss that uses the teacher of one modality to supervise the student of the other. To mitigate noise from uncertain predictions and unreliable geometry, we formulate a \emph{dual confidence weighting mechanism}.

For supervising the 2D student with the 3D teacher, we compute a weighted cross-entropy loss 
\begin{equation}
    \mathcal{L}_{C}^{2D} = - \sum_{j} w_i \cdot \log(S_{2D}^{y_i}(I_j)),
\end{equation}
for each pixel $j$ corresponding to a valid and visible 3D point $p_i$, where $y_i = \arg\max(T_{3D}(p_i))$ is the hard pseudo-label from the 3D teacher, $S_{2D}^{y_i}(I_j)$ is the 2D student's probability output for class $y_i$ at pixel $j$, and the weight
\begin{equation}
    w_i = \underbrace{\max(\text{softmax}(T_{3D}(p_i)))}_{\text{prediction confidence}} \cdot \underbrace{c_i^{\text{rec}}}_{\text{reconstruction confidence}}
\end{equation}
combines two confidence scores, \emph{prediction confidence} and \emph{reconstruction confidence}.
Here, $T_{3D}(p_i)$ is the 3D teacher's logit output, and $c_i^{\text{rec}}$ is MapAnything's per-point reconstruction confidence. This dual filtering ensures that supervision is dominated by reliable predictions on well-reconstructed geometry. A symmetric loss $\mathcal{L}_{C}^{3D}$ supervises the 3D student using the 2D teacher.

To prevent overconfident wrong predictions from dominating, we apply stronger augmentations (RandomCrop, Cutout, AugMix for 2D; RandomRotation, RandomScale, RandomJitter for 3D) to student inputs compared to teachers, encouraging the student to learn robust features while the teacher provides stable targets.

\subsection{Training Objective}
We combine our proposed cross-modal consistency loss with standard intra-modal supervised ($\mathcal{L}_{S}^{m}$) and unsupervised ($\mathcal{L}_{U}^{m}$) losses for each modality $m \in \{2D, 3D\}$. The final training objective $\mathcal{L}_{\text{Total}}$ is a weighted sum of all components:
\vspace{0.05cm}
\begin{equation}
\mathcal{L}_{\text{Total}} = \hspace{-0.2cm} \sum_{m \in \{2D, 3D\}}\hspace{-0.4cm} (\mathcal{L}_{S}^{m} + \mathcal{L}_{U}^{m}) \hspace{0.2cm} + \hspace{0.2cm}\lambda_{2D}\mathcal{L}_{C}^{2D} + \lambda_{3D}\mathcal{L}_{C}^{3D}
\end{equation}
\vspace{0.05cm}
where $\lambda_{2D}$ and $\lambda_{3D}$ are hyperparameters that balance the contribution of the cross-modal consistency terms.

\section{Experiments}
\label{sec:experiments}
We evaluate our method on four datasets spanning outdoor and indoor scenes: KITTI-360~\cite{liao_kitti-360_2023}, Waymo Open Dataset~\cite{sun_scalability_2020}, Cityscapes~\cite{cordts_cityscapes_2016} and NYUv2~\cite{couprie_indoor_2013}. Our experiments demonstrate that cross-modal consistency from reconstructed 3D geometry substantially outperforms traditional sparsely annotated segmentation methods.
\subsection{Experimental Setup}
\myparagraph{Datasets.}
\emph{KITTI-360} is a large-scale outdoor dataset with RGB images from the city of Karlsruhe, covering 19 semantic classes. The dataset provides accumulated LiDAR scans, which we use for baseline comparisons with ground-truth 3D geometry.
\emph{Waymo Open Dataset} provides images from diverse driving environments across multiple US cities, with 25 semantic classes. Unlike KITTI-360, it provides non-accumulated, single-scan LiDAR point clouds.
\emph{Cityscapes} is a widely-used urban driving dataset captured in 50 European cities, featuring 19 semantic classes. 
\emph{NYUv2} is an indoor RGB-D dataset with images across 40 semantic classes. We use the provided depth maps to generate 3D point clouds for baseline comparisons.
For all datasets, we generate scribble labels using Scribbles4All\,\cite{boettcher_scribbles_2024}, covering approx.~2-4\% of pixels while maintaining class distribution similar to full annotations. For Cityscapes, we also utilize the provided coarse labels and derive point annotations for a comprehensive comparison of sparse annotations. Specifically, we obtain point labels by randomly sampling one pixel per class in an image, following \cite{das2023weakly}. We treat Cityscapes as a collection of single images, as the training split consists of irregular drives, and thus perform reconstruction on a per-image basis.

\myparagraph{Implementation Details.}
We use SegFormer-B4~\cite{xie_segformer_2021} for 2D and Point Transformer V3~\cite{wu_point_2024} for 3D segmentation respectively. Training runs for 50 epochs (250 for NYUv2) with batch size 12 on two H100 GPUs. We use AdamW~\cite{loshchilov2017decoupled} optimizer with learning rates $5 \times 10^{-5}$ (2D) and $1 \times 10^{-3}$ (3D). We train our framework in two stages. Base Training (15 epochs for KITTI-360/Waymo, 150 for NYUv2) establishes independent student-teacher setups for each modality using sparse supervision. We then introduce the CMC loss and ramp it up linearly over 5 epochs to maximum weight $\lambda = 0.1$.
For data augmentation, student's images receive more severe augmentations~(Cutout, Blur, AugMix for 2D; RandomRotation, RandomScale, RandomJitter for 3D) while teacher images receive weaker ones, encouraging robust feature learning. We reconstruct scenes using MapAnything~\cite{keetha2025mapanything} with up to 200 images per batch, then apply our view-aware sampling to create 120K-point clouds (60\% from the current view, 40\% from surrounding context).

\begin{figure*}[htb]
    \centering
    \vspace{-0.2cm}
    \includegraphics[width=\linewidth]{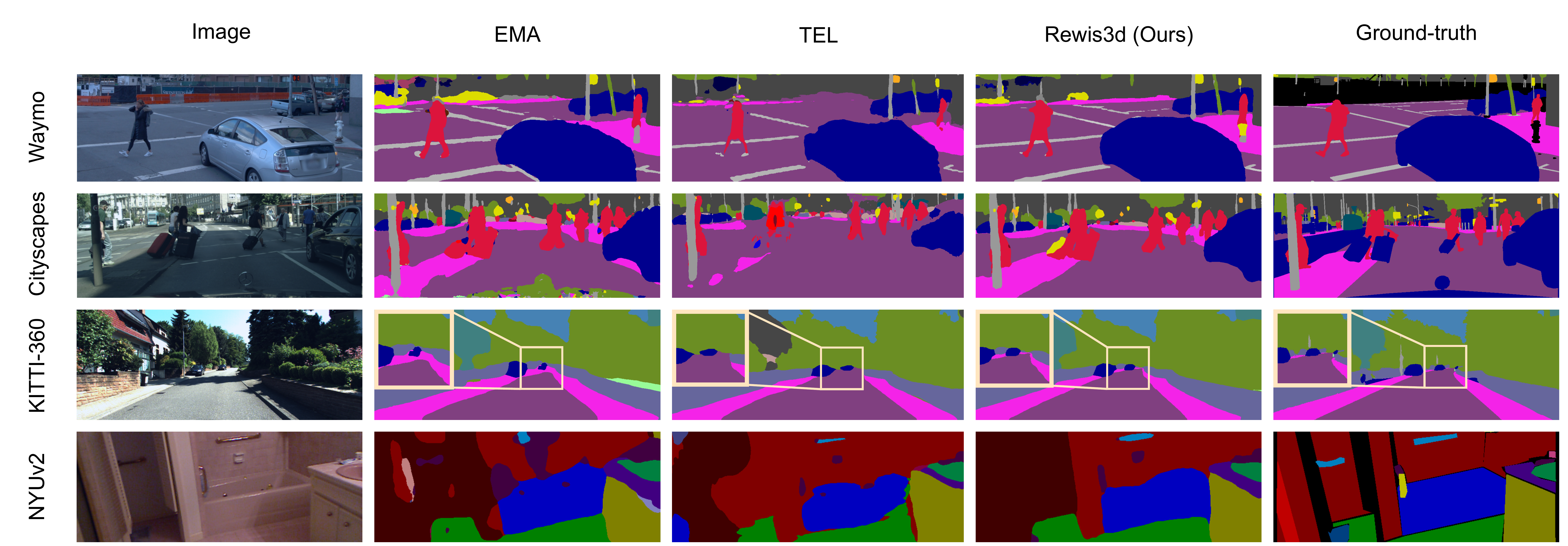}
    \vspace{-0.7cm}
    \caption{\textbf{Qualitative comparison across outdoor and indoor datasets.} \method produces sharper boundaries, more accurate fine-grained predictions, and better long-range segmentation compared to the Mean Teacher baseline (EMA) and TEL, even in regions where 3D reconstruction is uncertain. Colormaps are provided in the appendix.}
    \label{fig:qualitative}
    \vspace{-0.4cm}
\end{figure*}

\subsection{Baselines}
We compare against state-of-the-art methods specifically designed for sparsely annotated semantic segmentation, Tree Energy Loss (TEL)~\cite{liang_tree_2022} and SASFormer~\cite{su_sasformer_2023}. These methods represent current best practices for learning directly from sparse supervision without external priors. As a foundational baseline, we train SegFormer-B4 with a standard Mean Teacher setup on scribbles only (denoted as EMA), which serves as the basis for our CMC framework and demonstrates weakly-supervised consistency regularization without geometric guidance.

To isolate the contribution of 3D geometry, we evaluate two variants of \method. Ours (Recon) represents our main contribution, using point clouds reconstructed from multi-view stereo. Ours (Real 3D) uses ground-truth LiDAR data, where available, to serve as a reference for 3D geometry.

Finally, we include a fully supervised baseline with dense annotations to quantify how much of the supervision gap our method closes. All methods are trained according to their official implementations.

\subsection{Main Results}\label{sec:main_results}
\cref{tab:main_results_2d} presents our main 2D semantic segmentation results. Our method achieves substantial improvements across all datasets, demonstrating the effectiveness of cross-modal geometric supervision.

\myparagraph{Improvements over Specialized WSSS Methods.}
\method substantially outperforms existing weakly-supervised methods specifically designed for scribble supervision, with particularly pronounced gains on scene-centric outdoor datasets. This performance gap highlights the crucial role of geometric context in propagating weak labels. On Waymo, Ours (Recon) achieves 53.3\% mIoU compared to 49.4\% for EMA, 42.4\% for TEL, and 37.8\% for SASFormer, representing improvements of 3.9,10.9, and 15.5 points, respectively. The trend is consistent on KITTI-360, where we reach 63.9\% mIoU versus 60.3\% (EMA), 59.2\% (TEL), and 46.4\% (SASFormer).
This geometric advantage also extends to indoor environments, which often contain objects with less distinct 3D structure. On the NYUv2 dataset, \method (Recon) achieves 46.1\% mIoU, once again outperforming all scribble-supervised baselines, including the next-best SASFormer (44.7\%).

These results highlight a fundamental limitation of WSSS methods that operate solely in the 2D image plane: they struggle to robustly resolve ambiguities and propagate supervision across complex, occluded regions using only appearance cues. In contrast, our approach leverages 3D geometric information (derived from either single-view or multi-view reconstruction) to project weak labels into a stable 3D space. This geometric grounding provides a powerful mechanism for consistency enforcement and long-range supervision propagation, effectively enabling the sparse scribble to govern the segmentation of an entire 3D scene volume, rather than being confined to propagation within single frames.

\myparagraph{Reconstructed 3D Outperforms Real 3D.}
Interestingly, \method using reconstructed point clouds consistently outperforms the variant using ground-truth LiDAR/depth sensors. We argue that this counterintuitive result stems from two primary advantages of the reconstructed data. First, the reconstructions typically offer a denser and more complete point cloud representation compared to the sparsity inherent in real-world LiDAR or depth sensor scans, which is critical for robust geometric label propagation. Second, the reconstructed 3D allows us to leverage our Dual Confidence Filtering Approach, which assigns lower weights to uncertain points derived from the reconstruction process itself. In stark contrast, when using real LiDAR or depth sensor data (Real 3D), we do not have an estimation of the reconstruction confidence, forcing us to treat all real 3D points equally. This lack of a confidence measure prevents the robust filtering of noise and imprecise measurements inherent in raw sensor data, leading to less effective supervision propagation compared to the confidence-weighted reconstructed point clouds. Our ablation studies (Table~\ref{tab:ablation_comprehensive}) quantitatively confirm both the value of dense multi-view reconstruction and the effectiveness of dual confidence filtering.

\subsection{Generalization to Diverse Annotation Types}
To validate the general applicability of our framework beyond scribble-based supervision, we conduct experiments on the Cityscapes dataset using three distinct types of sparse annotations: points, scribbles, and coarse labels. Point labels represent the sparsest form of supervision, where each object instance is marked with a single pixel. Scribble labels are generated as on other datasets, and the provided coarse labels serve as an upper bound for sparse, yet inexact, supervision. Importantly, as Cityscapes is treated as a collection of single images, our 3D reconstruction is performed on a per-frame basis, demonstrating our method's utility even without multi-view video context and label accumulation.

\cref{tab:cityscapes_results} presents the results of this analysis. \method demonstrates significant performance gains over the EMA baseline across all three annotation types. With point supervision, our method improves the mIoU by 6.0\% (from 50.5\% to 56.5\%), demonstrating that even from minimal spatial cues, our 3D-to-2D consistency mechanism can effectively propagate semantic information. The performance gain is even larger for scribbles at 6.9\% (from 61.2\% to 68.1\%). For coarse annotations, which already provide a strong baseline, our method still achieves a consistent improvement of 2.1\% (from 66.5\% to 68.6\%). These findings generalize to KITTI-360 and Waymo in the case of points and scribbles as illustrated in Fig.\,\ref{fig:teaser}.

These insights underscore the robustness and versatility of our approach. The consistent improvements, regardless of the specific form of sparse annotation, confirm that leveraging reconstructed 3D geometry is a powerful and generalizable strategy for weakly-supervised semantic segmentation. This demonstrates that our method is not tailored to a single annotation type but rather provides a general improvement for a wide range of sparse supervision scenarios.

\begin{table}[tb]
    \centering
    \caption{
        \textbf{Generalization across annotation types on Cityscapes (mIoU \%).}
        \method consistently outperforms all competing methods, including the EMA baseline, across three sparse annotation types—points, scribbles, and coarse masks—demonstrating strong versatility and broad applicability. Reported improvements (+6.0, +6.9, +2.1) are relative to the EMA baseline, with the largest gains observed under minimal supervision (points and scribbles), confirming the effectiveness of geometric consistency in diverse weak supervision scenarios.
    }
    \label{tab:cityscapes_results}
    \vspace{-0.2cm}
    \fontsize{8pt}{10pt}\selectfont
    \begin{tabular}{l @{\hspace{0.9cm}} c c c}
        \toprule
        \multirow{2}{*}{\textbf{Method}} & \multicolumn{3}{c}{\textbf{Annotation Type}} \\
        
        & \emph{Points} & \emph{Scribbles} & \emph{Coarse} \\
        \midrule
        Fully Supervised & 77.6 & 77.6 & 77.6 \\
        \midrule
        TEL~\cite{liang_tree_2022} & \underline{53.1} & \underline{64.4} & 64.9 \\
        SASFormer~\cite{su_sasformer_2023} & 42.7 & 55.6 & 42.8 \\
        EMA (Baseline) & 50.5 & 61.2 & \underline{66.5} \\
        \textbf{Ours} & \textbf{56.5} & \textbf{68.1} & \textbf{68.6} \\
        \midrule
        \textbf{Improvement} & \textcolor{dgreen}{+6.0} & \textcolor{dgreen}{+6.9} & \textcolor{dgreen}{+2.1} \\
        \bottomrule
    \end{tabular}
    \vspace{-0.3cm}
\end{table}

\subsection{Qualitative Results}
\Cref{fig:qualitative} reveals dataset-specific qualitative strengths of our approach. On Waymo, \method significantly improves nearby object predictions, producing sharper class boundaries and more accurate classification. Fine-grained classes (e.g., road and lane markers) that challenge the Mean Teacher (EMA) baseline are predicted with substantially higher detail, as our geometric consistency constraints effectively resolve ambiguities in these complex regions.

For KITTI-360, \method demonstrates remarkable improvements in distant scene regions. Despite inherent 3D reconstruction uncertainty at long ranges, our framework successfully propagates semantic information, producing cleaner road-sidewalk boundaries and more coherent distant vehicles. This suggests the bidirectional consistency mechanism effectively leverages even noisy geometric cues to enhance segmentation.

On NYUv2, improvements reflect the dataset's unique challenges and 40 classes. Many classes lack strong 3D structure (e.g., towels), limiting the distinctiveness of our geometric signal compared to outdoor scenes. Nevertheless, \method still clearly improves over baselines: boundaries are more precisely delineated, and predictions show better spatial coherence, particularly for structured objects like furniture. This demonstrates that while most effective with pronounced 3D structure, our geometric supervision still provides measurable benefits and state-of-the-art performance in ambiguous indoor environments.

\begin{table*}[htb]
\centering
\caption{\textbf{Ablation study on Waymo (mIoU \%).} We systematically validate each component of our framework. Base configuration uses EMA baseline (49.4 mIoU). Each row progressively adds components, demonstrating their individual and cumulative contributions.}
\label{tab:ablation_comprehensive}
\vspace{-0.1cm}
\resizebox{\textwidth}{!}{\begin{tabular}{@{}lccccccc}
\toprule
\multirow{2}{*}{\textbf{Configuration}} & \multicolumn{3}{c}{\textbf{Confidence Filtering}} & \multicolumn{2}{c}{\textbf{Sampling}} & \multirow{2}{*}{\textbf{3D Source}} & \multirow{2}{*}{\textbf{mIoU}} \\
& None & Pred. & Recon. & Random & View-Aware & & \\ \midrule
EMA Baseline (2D only) & --- & --- & --- & --- & --- & --- & 49.4\\
\midrule
\multicolumn{1}{@{}l}{\emph{Effect of Confidence Filtering (with View-Aware sampling, Multi-view Recon.):}} \\
\quad No filtering & \checkmark & & & & \checkmark & Multi-view & 51.9 \\
\quad + Prediction conf. & & \checkmark & & & \checkmark & Multi-view & 52.7 \\
\quad + Reconstruction conf. & & & \checkmark & & \checkmark & Multi-view & 52.1 \\
\quad + Both (Ours) & & \checkmark & \checkmark & & \checkmark & Multi-view & \textbf{53.3} \\
\midrule
\multicolumn{1}{@{}l}{\emph{Effect of Sampling Strategy (with dual confidence, Multi-view Recon.):}} \\
\quad Random sampling & & \checkmark & \checkmark & \checkmark & & Multi-view & 51.9 \\
\quad View-Aware & & \checkmark & \checkmark & & \checkmark & Multi-view & \textbf{53.3} \\
\midrule
\multicolumn{1}{@{}l}{\emph{Effect of 3D Reconstruction Quality (with dual confidence, View-Aware):}} \\
\quad Single frame & & \checkmark & \checkmark & & \checkmark & Single & 52.1 \\
\quad Multi-view (Ours) & & \checkmark & \checkmark & & \checkmark & Multi-view & \textbf{53.3} \\
\bottomrule
\end{tabular}}
\vspace{-0.4cm}
\end{table*}

\subsection{Ablation Studies}
\label{sec:ablations}
We conduct comprehensive ablation studies to validate our key design choices. All experiments are performed on Waymo except for scribble length ablations, for which we use KITTI-360. Results are summarized in \cref{tab:ablation_comprehensive}.
\begin{figure}[tb]
\centering
\vspace{0.2cm}
\includegraphics[width=1.0\linewidth]{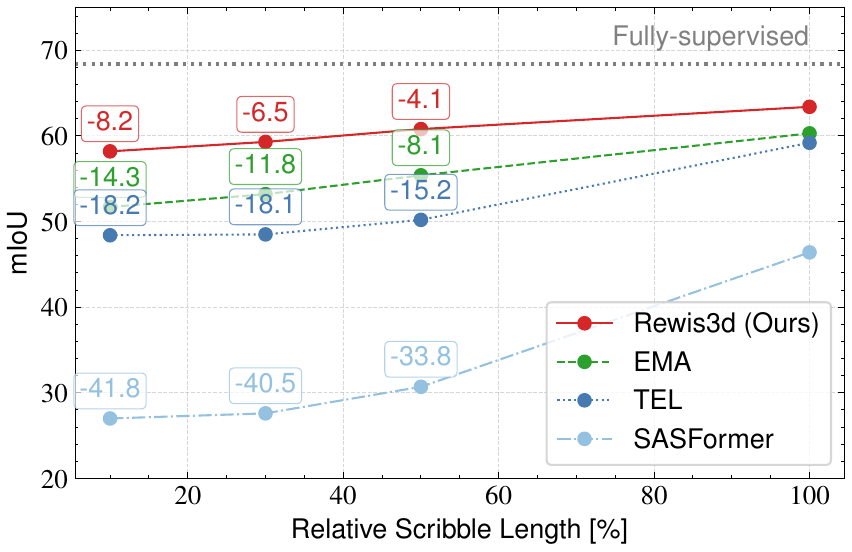}
\vspace{-0.6cm}
\caption{\textbf{Scribble length ablation on KITTI-360.} \method~(red) maintains strong performance across varying scribble lengths compared to baselines, with largest gains in sparse settings, demonstrating the value of geometric supervision when annotations are scarce.}
\label{fig:scribble_plot}
\vspace{-0.3cm}
\end{figure}

\myparagraph{Key Findings from Ablations.}
\textbf{(1) Dual confidence filtering improves reliability:} Both prediction confidence (+0.8 mIoU) and reconstruction confidence (+0.2 mIoU) contribute to filtering noisy pseudo-labels. Without any filtering, performance reaches 51.9 mIoU. Their combination achieves the best result (53.3 mIoU), showing they capture complementary aspects of reliability.

\textbf{(2) View-aware sampling enhances correspondence quality:} Random sampling yields only $\sim$140 correspondences per image, limiting cross-modal learning (51.9 mIoU). Our view-aware strategy improves performance by +1.4 mIoU, achieving 53.3 mIoU.

\textbf{(3) Multi-view reconstruction provides richer geometry:} Dense point clouds from multi-view video sequences (53.3 mIoU) outperform single-frame reconstruction (52.1 mIoU) by +1.2 mIoU, as they provide richer geometric context and more reliable depth estimates through multi-view consistency. However, single-frame reconstruction still improves over the baseline by +2.7 mIoU, enabling application even in datasets without video data.

\myparagraph{Scribble Length Robustness.}
\cref{fig:scribble_plot} demonstrates that our framework maintains strong performance even with extremely sparse scribbles. While the EMA baseline and other methods degrade significantly as annotation density decreases, our method remains robust, with the performance gap widening in sparser regimes—precisely where annotation efficiency matters most.

\myparagraph{Orthogonality to Backbone Architecture.}
 Our proposed framework is architecture-agnostic and functions independently of the underlying segmentation backbone. To validate this, we replaced our standard network with the Encoder-only Mask Transformer (EoMT)~\cite{kerssies2025your}, which relies on semantic features distilled from DINOv2~\cite{oquab_dinov2_2024}. As shown in App.\,\ref{supp:quantitative}, integrating Rewis3d consistently elevates performance over the vanilla EoMT baseline across all scenarios, demonstrating that our geometric priors successfully complement semantic foundational features.

\section{Conclusion}
\label{sec:conclusion}
We introduced \method, which successfully leverages 3D geometry from reconstructed point clouds to significantly improve sparsely-supervised semantic segmentation across multiple supervision types, closing a substantial portion of the performance gap to fully supervised models. Our work demonstrates that for complex, scene-centric data, targeted geometric consistency provides an effective supervisory signal. By generating this 3D supervision from standard 2D videos, our framework obviates the need for specialized 3D sensors, making it broadly applicable.

\myparagraph{Limitations and Future Work.}
Our framework achieves state-of-the-art results by leveraging a 3D reconstruction model that is not explicitly optimized for dynamic content. On sequential driving data, this can introduce geometric noise and depth uncertainties from unmodeled dynamic objects or distant regions. The robustness of \method, which succeeds despite these geometric artifacts, underscores the strength of our core cross-modal consistency (CMC) principle. This also defines a clear and compelling direction for future research: integrating reconstruction models that explicitly handle dynamic scenes. We hypothesize that providing the CMC loss with a temporally consistent and geometrically cleaner 3D signal would further strengthen supervision, unlocking substantial new performance gains.

\section*{Acknowledgements}
Jan Eric Lenssen is supported by the German Research Foundation (DFG) - 556415750 (Emmy Noether Programme, project: Spatial Modeling and Reasoning).
{
    \small
    \bibliographystyle{ieeenat_fullname}
    \bibliography{main}
}
\appendix

\renewcommand\thesection{\Alph{section}}
\numberwithin{equation}{section}
\numberwithin{figure}{section}
\numberwithin{table}{section}
\renewcommand{\thefigure}{\thesection\arabic{figure}}
\renewcommand{\thetable}{\thesection\arabic{table}}
\crefname{appendix}{Sec.}{Secs.}

{\onecolumn 
{\begin{center}
\Large\bf
{Rewis3d: Reconstruction for Weakly-Supervised Semantic Segmentation}\\[1em]
\large
Supplementary Material
\end{center}
}
\newcommand{\additem}[2]{\item[\textbf{(\ref{#1})}] \textbf{#2} \dotfill\makebox{\textbf{\pageref{#1}}}}

\newcommand{\myindent}{.5em}
\newcommand{\addsubitem}[2]{\vspace{.5em}\textbf{(\ref{#1})}\hspace{\myindent} #2 \\}

\newcommand{\adddescription}[1]{\vspace{.1em}\begin{adjustwidth}{0cm}{0cm}#1\end{adjustwidth}}
\setlist[itemize]{noitemsep,leftmargin=*,topsep=0em}
\setlist[enumerate]{noitemsep,leftmargin=*,topsep=0em}

\noindent In this supplement to our work on \method -- Reconstruction for Weakly-Supervised Semantic Segmentation, we provide further results, derivations, and implementation details, as indexed below. We \textbf{particularly encourage} the reader to see the added information on 3D segmentation performance in App.~\ref{supp:3d-seg} and the additional discussion on the merits of reconstructed 3D over LiDAR in App.~\ref{supp:real-vs-recon}.
\vspace{0.3in}

\begin{adjustwidth}{0.15cm}{0.15cm}
\begin{enumerate}[label={({\arabic*})}, topsep=1em, itemsep=.2em]
    \additem{supp:3d-seg}{\method Improves 3D Segmentation}\\[0.4em]\hfill
    \additem{supp:qualitative}{Further Qualitative Results}\\[0.4em]\hfill
    \additem{supp:quantitative}{Quantitative Results}\\[0.4em]\hfill
    \additem{supp:CMCsampling}{Sampling Strategies for Reconstruction}\\[0.4em]\hfill
    \additem{supp:real-vs-recon}{Analysis of Real vs. Reconstructed Supervision}\\[0.4em]\hfill
    \additem{supp:data-gen}{Label Generation}\\[0.4em]\hfill
    \additem{supp:implementation}{Implementation Details \& Hyperparameters}\\[0.4em]\hfill
    \additem{supp:colormap}{Colormaps}\\[0.4em]
\end{enumerate}
\end{adjustwidth}
}
\setlength{\parskip}{.5em}
\clearpage

\section{\method Improves 3D Segmentation}\label{supp:3d-seg}
\begin{figure*}[h]
    \centering
    \includegraphics[width=\linewidth]{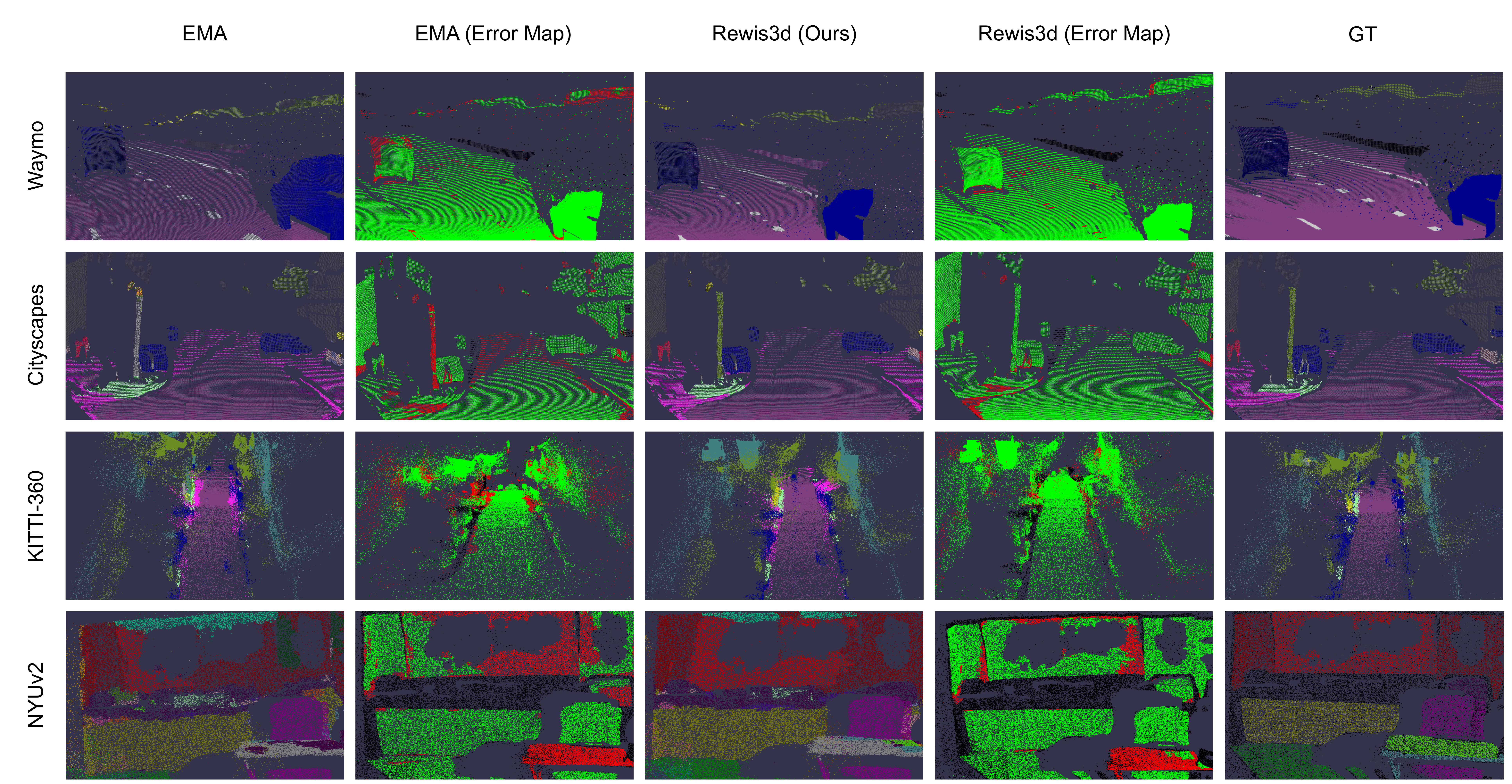}
    \caption{\textbf{3D Segmentation and Error Maps.} \method improves also improves the 3D segmentation noticeably. The bidirectional knowledge transfer through CMC especially improves the 3D segmentation quality with respect to errors emanating from misclassifications of objects. This highlights how segmentation models in 2D and 3D domain possess different advantages. While the structure of the 3D space lends itself to more accurate separation of objects, the correct class assignment appears to be easier to learn in 2D.}
    \label{fig:qualitative_3d}
\end{figure*}

As shown in Table~\ref{tab:main_results_3d}, cross-modal learning significantly benefits 3D segmentation. Our CMC framework improves 3D mIoU by +3.7 on Waymo and +4.1 on NYUv2 compared to the 3D-only EMA baseline, demonstrating effective bidirectional knowledge transfer. Due to the lack of ground truth labels for the reconstructed point clouds, we evaluate against unprojected 2D segmentation masks. While this introduces some uncertainty in the reference labels, the validity of relative performance comparisons remains intact. Thus, we demonstrate reliable improvements in the 3D modality, which are visually corroborated in Fig.~\ref{fig:qualitative_3d}.

\begin{table*}[h]
    \centering
    \vspace{0.5cm}
    \caption{
        \textbf{3D Semantic Segmentation Performance (mIoU \%).} Cross-modal consistency also improves 3D segmentation through bidirectional knowledge transfer from 2D. The column {$\Delta$ vs EMA} denotes the absolute performance improvement in percentage points\,(pp) of our method compared to the 3D-only Mean Teacher baseline (EMA).
    }
    \label{tab:main_results_3d}
    \footnotesize
    \begin{tabular}{lS[table-format=2.1] S[table-format=2.1] S[table-format=2.1] S[table-format=2.1] S[table-format=2.1] S[table-format=2.1]}
        \toprule
        \multirow{2}{*}{\textbf{Method}} & \multicolumn{2}{c}{\textbf{Waymo}} & \multicolumn{2}{c}{\textbf{KITTI-360}} & \multicolumn{2}{c}{\textbf{NYUv2}} \\
        \cmidrule(lr){2-3} \cmidrule(lr){4-5} \cmidrule(lr){6-7}
        & {\textbf{mIoU}} & {$\Delta$ vs EMA} & {\textbf{mIoU}} & {$\Delta$ vs EMA} & {\textbf{mIoU}} & {$\Delta$ vs EMA} \\
        \midrule
        EMA (3D only) & 41.8 & \textendash & 44.3 & \textendash & 24.4 & \textendash\\
        Ours (Recon) & \textbf{45.5} & \textbf{+3.7} & \textbf{44.9} & \textbf{+0.6} & \textbf{28.5} & \textbf{+4.1}\\
        \bottomrule
    \end{tabular}
    \vspace{1cm}
\end{table*}

\clearpage
\section{Further Qualitative Results}\label{supp:qualitative}

\begin{figure*}[h]
    \centering
    \includegraphics[width=\linewidth]{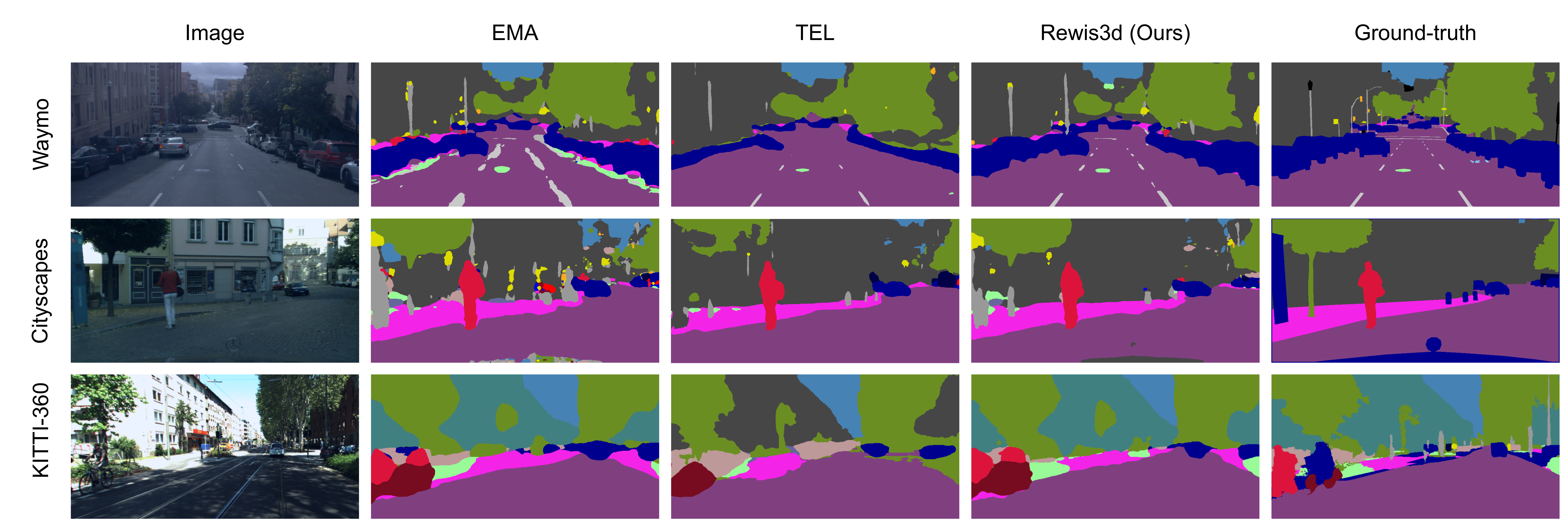}
    \caption{\textbf{Qualitative comparison with point supervision.} Even with minimal supervision (one point per object), \method successfully propagates labels to the full object extent. Note the improved segmentation of the vehicle and road markers in Waymo (top row) and the clearer delineation of the sidewalk in KITTI-360 (bottom row) compared to the EMA and TEL baselines. Additionally, we observe that the KITTI-360 ground truth contains occasional labeling errors (e.g., the bicycle); consequently, our model's visually correct predictions in these regions may be penalized during quantitative evaluation against the imperfect ground truth.}
    \label{fig:qualitative_point}
\end{figure*}

\begin{figure*}[h]
    \centering
    \includegraphics[width=\linewidth]{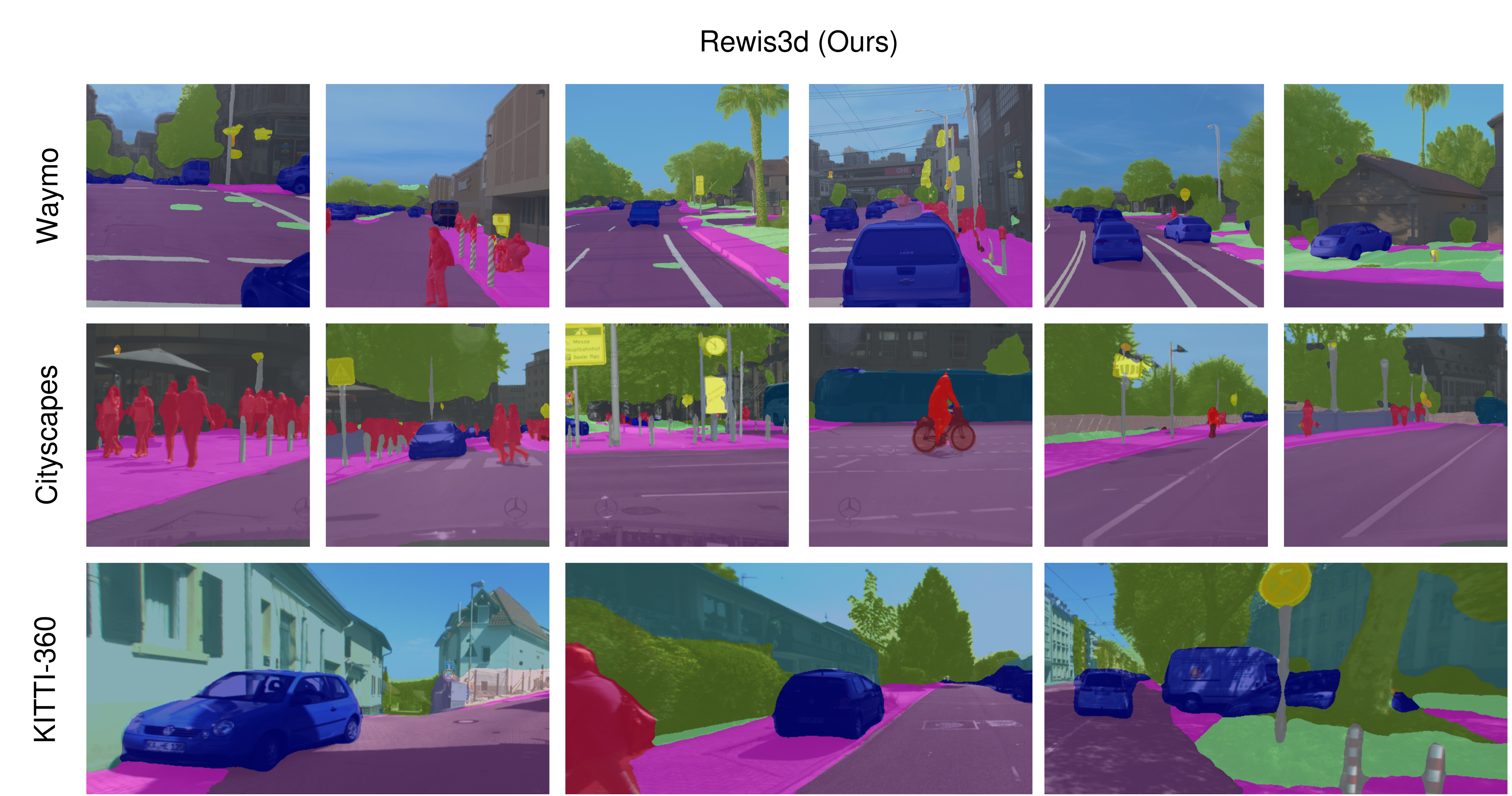}
    \caption{\textbf{Additional Qualitative Results with Scribble Supervision.} We provide further qualitative examples of \method predictions on the Waymo (top), Cityscapes (middle), and KITTI-360 (bottom) datasets. These visualizations demonstrate the model's capability to generate spatially coherent segmentations with precise boundaries across diverse outdoor urban environments, solely trained with sparse scribble supervision.}
    \label{fig:qualitative_scribble}
\end{figure*}
\clearpage
\begin{landscape}
\section{Quantitative Results}\label{supp:quantitative}
This section presents detailed class-wise evaluations to complement the experiments discussed in the main paper. Table~\ref{tab:waymo_comparison_final} details the individual performance of \method on Waymo using scribble and point labels. Similarly, comprehensive results for NYUv2 and KITTI-360 are provided in Table~\ref{tab:nyuv2_comparison_final} and Table~\ref{tab:kitti360_cmc}, respectively. Finally, Table~\ref{tab:cityscapes_comparison} compares the class-wise performance of \method against competing methods across the three types of weak labels available for Cityscapes. The tables for the three outdoor datasets further contain results of using EoMT as the segmentation model with Rewis3d to demonstrate that our method is orthogonal to the improvements gained from relying on the strong foundational priors that EoMT leverages on its own.

\begin{table}[h]
    \hspace{1cm}
    \centering
    \caption[Comparison of 2D mIoU Results and Relative Scores on Waymo Dataset]{\textbf{Comparison of 2D mIoU Results and Relative Scores on Waymo Dataset} Bold numbers indicate the best and underlined values the second-best performance among non-fully supervised approaches.}
    \label{tab:waymo_comparison_final}
    \resizebox{1.3\textwidth}{!}{\begin{tabular}{l|cc|*{25}{c}}
        \toprule
        \textbf{Method} & \textbf{mIoU} & \textbf{SS/FS (\%)} & \rotatebox{90}{\textbf{bicycle}} & \rotatebox{90}{\textbf{bird}} & \rotatebox{90}{\textbf{building}} & \rotatebox{90}{\textbf{bus}} & \rotatebox{90}{\textbf{car}} & \rotatebox{90}{\textbf{construction cone pole}} & \rotatebox{90}{\textbf{cyclist}} & \rotatebox{90}{\textbf{ground}} & \rotatebox{90}{\textbf{ground animal}} & \rotatebox{90}{\textbf{lane marker}} & \rotatebox{90}{\textbf{motorcycle}} & \rotatebox{90}{\textbf{motorcyclist}} & \rotatebox{90}{\textbf{other large vehicle}} & \rotatebox{90}{\textbf{pedestrian}} & \rotatebox{90}{\textbf{pedestrian object}} & \rotatebox{90}{\textbf{pole}} & \rotatebox{90}{\textbf{road}} & \rotatebox{90}{\textbf{road marker}} & \rotatebox{90}{\textbf{sidewalk}} & \rotatebox{90}{\textbf{sign}} & \rotatebox{90}{\textbf{sky}} & \rotatebox{90}{\textbf{traffic light}} & \rotatebox{90}{\textbf{trailer}} & \rotatebox{90}{\textbf{truck}} & \rotatebox{90}{\textbf{vegetation}} \\
        \midrule
        Fully Supervised & 59.1 & - & 79.3 & 10.0 & 93.4 & 16.0 & 93.9 & 65.4 & 57.9 & 74.6 & 0.0 & 64.3 & 73.7 & 0.0 & 16.0 & 83.0 & 9.0 & 69.7 & 94.8 & 64.9 & 84.3 & 73.7 & 95.3 & 74.0 & 28.1 & 66.3 & 89.2 \\
        \midrule
        \multicolumn{1}{@{}l}{\emph{Point annotations}} \\
        EMA & \underline{43.4} & \underline{73.4} & \underline{58.1} & \textbf{0.0} & \underline{84.8} & 15.7 & \underline{80.7} & \underline{39.6} & 10.9 & \underline{50.3} & \textbf{0.0} & 30.9 & \underline{54.0} & \textbf{0.0} & \textbf{10.8} & \underline{58.7} & \underline{7.0} & \underline{43.9} & 85.3 & \underline{43.1} & \underline{60.8} & \underline{49.4} & 92.0 & \underline{54.3} & \underline{22.4} & \underline{54.3} & \underline{78.8} \\
        SASFormer & 37.4 & 63.2 & 52.9 & \textbf{0.0} & 80.7 & \underline{{16.4}} & 75.5 & 11.9 & \underline{15.4} & 49.8 & \textbf{0.0} & 32.7 & 47.1 & \textbf{0.0} & 0.1 & 52.9 & \underline{7.0} & 23.9 & \underline{89.9} & 29.9 & 57.4 & 44.0 & \textbf{93.8} & 32.0 & 1.7 & 40.8 & 78.3 \\
        TEL & 35.4 & 59.9 & 0.0 & \textbf{0.0} & 78.5 & \textbf{{41.9}} & 54.3 & 21.2 & 9.4 & 45.7 & \textbf{0.0} & \textbf{53.9} & 28.5 & \textbf{0.0} & 2.3 & 49.3 & 0.0 & 34.8 & 86.4 & 39.7 & 55.1 & 45.4 & \underline{92.9} & 41.2 & 1.9 & 29.8 & 73.3 \\
        Ours (Recon) & \textbf{48.8} & \textbf{82.6} & \textbf{59.9} & \textbf{0.0} & \textbf{88.0} & 6.1 & \textbf{89.5} & \textbf{44.3} & \textbf{23.7} & \textbf{64.1} & \textbf{0.0} & \underline{53.6} & \textbf{60.2} & \textbf{0.0} & \underline{7.6} & \textbf{66.5} & \textbf{7.3} & \textbf{49.9} & \textbf{91.9} & \textbf{54.4} & \textbf{72.9} & \textbf{57.7} & 91.0 & \textbf{55.9} & \textbf{{28.8}} & \textbf{64.6} & \textbf{81.2} \\
        \midrule 
        \multicolumn{1}{@{}l}{\emph{Scribble annotations}} \\
        EMA & \underline{49.4} & \underline{83.6} & \underline{71.9} & \textbf{0.0} & 87.8 & 15.4 & 84.8 & \underline{46.9} & \underline{42.9} & 59.6 & \underline{0.0} & 39.1 & \underline{65.4} & \textbf{0.0} & \underline{14.7} & 68.7 & 5.6 & \underline{52.7} & 89.1 & \underline{47.3} & 65.6 & \underline{56.3} & 92.8 & \underline{56.6} & 35.0 & \underline{57.8} & \underline{81.6} \\
        SASFormer & 37.8 & 64.0 & 48.8 & \textbf{0.0} & 81.1 & 23.4 & 81.5 & 12.8 & 8.4 & 51.7 & \underline{0.0} & 34.3 & 40.6 & \textbf{0.0} & 0.0 & 55.2 & 3.8 & 25.0 & \underline{91.1} & 32.3 & 58.5 & 44.5 & \textbf{93.7} & 33.2 & 0.9 & 46.1 & 78.2 \\
        TEL & 42.4 & 71.8 & 53.1 & \textbf{0.0} & 82.4 & \textbf{35.9} & 67.3 & 23.1 & 29.4 & 52.0 & \underline{0.0} & \underline{53.2} & 46.9 & \textbf{0.0} & 0.0 & 63.1 & 2.8 & 41.6 & 88.3 & 37.7 & 61.6 & 54.5 & \underline{93.4} & 50.9 & 1.0 & 45.8 & 76.4 \\
        Ours (Real 3D) & 47.8 & 80.9 & 62.2 & \textbf{0.0} & \underline{88.5} & 14.2 & \underline{87.5} & 35.9 & \textbf{44.2} & \underline{64.6} & \underline{0.0} & 33.2 & 63.2 & \textbf{0.0} & 5.7 & \underline{70.8} & \textbf{9.1} & 48.3 & 89.1 & 39.1 & \underline{71.2} & 52.3 & 90.1 & 45.9 & \textbf{46.3} & 52.7 & 81.3 \\
        Ours (Recon) & \textbf{53.3} & \textbf{90.2} & \textbf{72.2} & \textbf{0.0} & \textbf{88.9} & \underline{30.1} & \textbf{90.0} & \textbf{51.5} & 37.5 & \textbf{68.3} & \textbf{0.1} & \textbf{55.4} & \textbf{67.2} & \textbf{0.0} & \textbf{17.5} & \textbf{73.6} & \underline{7.5} & \textbf{55.5} & \textbf{92.7} & \textbf{58.3} & \textbf{76.0} & \textbf{62.8} & 91.0 & \textbf{59.0} & \underline{42.1} & \textbf{60.2} & \textbf{82.0} \\
        \midrule
        \multicolumn{1}{@{}l}{\emph{Scribble annotations \& EoMT backbone}} \\
        EMA & 53.13 & - & 65.44 & 0.00 & \textbf{88.77} & 15.97 & \textbf{90.19} & \textbf{51.47} & 55.24 & \textbf{62.41} & \textbf{0.71} & \textbf{43.00} & 60.37 & 0.00 & 10.41 & \textbf{75.11} & \textbf{27.79} & \textbf{57.93} & \textbf{90.74} & \textbf{48.33} & 72.10 & \textbf{62.61} & \textbf{91.94} & \textbf{58.49} & 45.03 & 73.68 & \textbf{80.64} \\
        Ours (Recon) & \textbf{54.04} & - & \textbf{67.18} & \textbf{5.19} & 88.05 & \textbf{32.40} & 90.15 & 45.22 & \textbf{65.26} & 61.93 & 0.00 & 39.75 & \textbf{62.83} & 0.00 & \textbf{21.85} & 70.34 & 25.58 & 57.67 & 89.58 & 44.90 & \textbf{72.27} & 60.24 & 90.09 & 57.14 & \textbf{45.70} & \textbf{77.89} & 79.81 \\
        \bottomrule
    \end{tabular}}
\end{table}

\begin{table}[h]
    \centering
    \caption[Detailed Class-wise mIoU Comparison on NYUv2 Dataset]{\textbf{Detailed Class-wise IoU Comparison on NYUv2 Dataset} Bold numbers indicate the best and underlined values the second-best performance among non-fully supervised approaches.}
    \label{tab:nyuv2_comparison_final}
    \resizebox{\linewidth}{!}{\begin{tabular}{l|cc|*{40}{c}}
        \toprule
        \textbf{Method} & \textbf{mIoU} & \textbf{SS/FS (\%)} & \rotatebox{90}{\textbf{background}} & \rotatebox{90}{\textbf{bag}} & \rotatebox{90}{\textbf{bathtub}} & \rotatebox{90}{\textbf{bed}} & \rotatebox{90}{\textbf{blinds}} & \rotatebox{90}{\textbf{books}} & \rotatebox{90}{\textbf{bookshelf}} & \rotatebox{90}{\textbf{box}} & \rotatebox{90}{\textbf{cabinet}} & \rotatebox{90}{\textbf{ceiling}} & \rotatebox{90}{\textbf{chair}} & \rotatebox{90}{\textbf{clothes}} & \rotatebox{90}{\textbf{counter}} & \rotatebox{90}{\textbf{curtain}} & \rotatebox{90}{\textbf{desk}} & \rotatebox{90}{\textbf{door}} & \rotatebox{90}{\textbf{dresser}} & \rotatebox{90}{\textbf{floor}} & \rotatebox{90}{\textbf{floor mat}} & \rotatebox{90}{\textbf{lamp}} & \rotatebox{90}{\textbf{mirror}} & \rotatebox{90}{\textbf{nightstand}} & \rotatebox{90}{\textbf{otherfurniture}} & \rotatebox{90}{\textbf{otherprop}} & \rotatebox{90}{\textbf{otherstructure}} & \rotatebox{90}{\textbf{paper}} & \rotatebox{90}{\textbf{person}} & \rotatebox{90}{\textbf{picture}} & \rotatebox{90}{\textbf{pillow}} & \rotatebox{90}{\textbf{refrigerator}} & \rotatebox{90}{\textbf{shelves}} & \rotatebox{90}{\textbf{sink}} & \rotatebox{90}{\textbf{sofa}} & \rotatebox{90}{\textbf{table}} & \rotatebox{90}{\textbf{television}} & \rotatebox{90}{\textbf{toilet}} & \rotatebox{90}{\textbf{towel}} & \rotatebox{90}{\textbf{wall}} & \rotatebox{90}{\textbf{whiteboard}} & \rotatebox{90}{\textbf{window}} \\
        \midrule
        Fully Supervised & 51.1 & - & 80.4 & 12.5 & 48.0 & 63.8 & 51.0 & 35.4 & 68.3 & 19.0 & 71.7 & 64.1 & 54.4 & 22.8 & 61.4 & 62.3 & 25.0 & 49.2 & 55.8 & 64.5 & 27.8 & 45.5 & 47.8 & 51.5 & 20.6 & 41.4 & 37.2 & 38.3 & 81.8 & 66.7 & 45.9 & 64.1 & 11.2 & 61.1 & 45.2 & 43.0 & 68.4 & 80.7 & 46.0 & 85.6 & 78.4 & 46.1 \\
        \midrule
        \multicolumn{1}{@{}l}{\emph{Scribble annotations}} \\
        EMA & 42.9 & 84.0 & 66.6 & 11.0 & 39.3 & 52.8 & 31.0 & 31.9 & 53.6 & 11.8 & 63.8 & 47.7 & 49.0 & \underline{21.9} & 59.0 & 47.1 & \textbf{20.5} & \textbf{48.5} & \underline{45.0} & 57.3 & \textbf{34.3} & 31.3 & 36.1 & 42.8 & 15.1 & 31.7 & 28.6 & 31.3 & 66.5 & 56.7 & 36.4 & 53.3 & 11.5 & 50.0 & \underline{38.6} & 36.8 & 57.9 & 69.5 & 36.3 & 80.2 & \underline{77.1} & 37.0 \\
        SASFormer & \underline{45.2} & \underline{88.5} & 71.2 & \underline{15.6} & \textbf{{50.8}} & \underline{59.3} & \underline{40.3} & \underline{34.0} & 56.0 & 13.6 & \textbf{69.0} & 51.8 & \underline{50.0} & \textbf{23.2} & 59.5 & 52.1 & 18.0 & 42.5 & 40.8 & 56.9 & \underline{32.5} & 34.8 & 34.3 & 38.7 & 14.3 & \underline{34.2} & 29.8 & 33.0 & \textbf{75.6} & 53.4 & \underline{40.3} & \textbf{59.7} & \textbf{12.4} & \textbf{57.5} & 36.9 & 35.5 & \textbf{61.4} & \underline{69.9} & \underline{40.1} & 79.2 & 72.8 & 38.4 \\
        TEL & 39.1 & 76.5 & 69.0 & \textbf{16.3} & 28.9 & 47.2 & 18.6 & 30.4 & 56.7 & 10.4 & 57.5 & 49.3 & 39.7 & 16.0 & 53.4 & 36.1 & 10.7 & 40.7 & 33.7 & 54.2 & 15.8 & 33.5 & 32.0 & 28.4 & \textbf{28.3} & 33.6 & 16.3 & \underline{35.4} & 69.9 & 51.1 & 35.3 & 51.6 & 6.7 & 47.5 & 31.3 & 27.3 & 41.5 & 63.4 & 25.8 & 76.6 & 74.4 & 38.6 \\
        Ours (Real 3D) & 44.7 & 87.6 & \underline{72.7} & 9.9 & \underline{47.4} & 55.9 & 39.7 & 31.4 & \underline{57.3} & \underline{14.0} & 62.8 & \underline{54.1} & 45.6 & 21.6 & \textbf{60.7} & \textbf{58.3} & 17.7 & 45.9 & \textbf{50.5} & \textbf{61.8} & 28.1 & \underline{35.5} & \textbf{41.3} & \underline{44.1} & 14.4 & 34.1 & \textbf{34.3} & 33.8 & 73.9 & \underline{59.7} & 36.4 & 54.1 & 9.8 & \underline{55.7} & 37.4 & \textbf{42.9} & 51.6 & 68.3 & 40.0 & \underline{80.9} & 70.1 & \underline{38.8} \\
        Ours (Recon) & \textbf{46.1} & \textbf{90.2} & \textbf{73.3} & 9.1 & 40.1 & \textbf{60.2} & \textbf{44.9} & \textbf{36.3} & \textbf{63.1} & \textbf{14.2} & \underline{67.2} & \textbf{58.9} & \textbf{51.0} & \underline{21.9} & \underline{60.6} & \underline{54.5} & \underline{18.5} & \underline{47.7} & 41.4 & \underline{61.6} & 32.2 & \textbf{43.5} & \underline{37.3} & \textbf{45.0} & \underline{16.1} & \textbf{38.4} & \underline{33.7} & \textbf{35.6} & \underline{74.3} & \textbf{65.2} & \textbf{{46.5}} & \underline{56.5} & \underline{{11.7}} & 55.3 & \textbf{41.6} & \underline{41.6} & \underline{60.1} & \textbf{74.7} & \textbf{43.6} & \textbf{82.2} & \textbf{77.5} & \textbf{44.1} \\
        \bottomrule
    \end{tabular}}
    \hspace{1cm}
\end{table}
\end{landscape}

\begin{table}
    \centering
    \hspace{1cm}
    \setlength{\tabcolsep}{5.5pt}
    \caption[Comparison of mIoU Results and Relative Scores on KITTI-360 Dataset]{\textbf{Comparison of mIoU Results and Relative Scores on KITTI-360 Dataset} Bold numbers indicate the best and underlined values the second-best performance among non-fully supervised approaches.}
    \label{tab:kitti360_cmc}
    \resizebox{0.85\linewidth}{!}{\begin{tabular}{l|cc|*{16}{c}}
        \toprule
        \textbf{Method} & \textbf{mIoU} & \textbf{SS/FS\,(\%)} & \rotatebox{90}{\textbf{bicycle}} & \rotatebox{90}{\textbf{building}} & \rotatebox{90}{\textbf{car}} & \rotatebox{90}{\textbf{fence}} & \rotatebox{90}{\textbf{motorcycle}} & \rotatebox{90}{\textbf{person}} & \rotatebox{90}{\textbf{pole}} & \rotatebox{90}{\textbf{road}} & \rotatebox{90}{\textbf{sidewalk}} & \rotatebox{90}{\textbf{sky}} & \rotatebox{90}{\textbf{terrain}} & \rotatebox{90}{\textbf{traffic light}} & \rotatebox{90}{\textbf{traffic sign}} & \rotatebox{90}{\textbf{truck}} & \rotatebox{90}{\textbf{vegetation}} & \rotatebox{90}{\textbf{wall}} \\
        \midrule
        Fully Supervised & 68.4 & - & 49.1 & 89.0 & 94.5 & 53.4 & 60.6 & 66.3 & 43.4 & 96.4 & 85.5 & 94.3 & 77.7 & 0.0 & 47.9 & 78.6 & 89.8 & 67.8 \\
        \midrule
        \multicolumn{1}{@{}l}{\emph{Point annotations}} \\
        EMA & \underline{52.2} & \underline{76.3} & \underline{33.1} & 78.2 & \underline{78.6} & \underline{42.0} & \underline{38.7} & \underline{43.3} & \underline{22.8} & \underline{84.4} & \underline{59.9} & \underline{84.7} & 56.7 & \textbf{0.0} & 33.6 & \underline{51.5} & \underline{80.0} & \underline{48.1} \\
        SASFormer & 27.0 & 39.5 & 21.0 & 54.6 & 22.2 & 28.2 & 3.8 & 6.7 & 0.1 & 65.6 & 27.5 & 53.2 & 26.0 & \textbf{0.0} & 8.2 & 23.5 & 60.9 & 30.5 \\
        TEL & 48.9 & 71.5 & 24.8 & \underline{79.4} & 70.6 & 39.5 & 37.0 & 37.9 & 16.0 & 81.1 & 54.1 & \textbf{86.9} & \underline{56.8} & \textbf{0.0} & \underline{34.8} & 42.9 & 76.3 & 43.7 \\
        Ours (Recon) & \textbf{58.2} & \textbf{85.1} & \textbf{43.7} & \textbf{83.9} & \textbf{89.7} & \textbf{49.5} & \textbf{48.8} & \textbf{46.4} & \textbf{25.3} & \textbf{91.0} & \textbf{70.8} & 72.2 & \textbf{61.8} & \textbf{0.0} & \textbf{37.9} & \textbf{63.9} & \textbf{83.8} & \textbf{63.1} \\
        \midrule
        \multicolumn{1}{@{}l}{\emph{Scribble annotations}} \\
        EMA & 60.3 & 88.1 & 43.4 & 83.5 & 86.7 & 46.5 & \textbf{59.7} & \underline{58.2} & 35.3 & 85.8 & 65.4 & 87.3 & 57.6 & \textbf{0.0} & 39.7 & \textbf{74.2} & 82.4 & 58.5 \\
        SASFormer & 46.4 & 67.8 & 35.8 & 72.7 & 78.3 & 42.1 & 23.3 & 21.7 & 4.5 & 74.4 & 52.1 & 76.9 & 61.4 & \textbf{0.0} & 17.2 & 55.6 & 78.9 & 47.8 \\
        TEL & 59.2 & 86.5 & \textbf{{50.5}} & 82.6 & 85.9 & 45.7 & 51.6 & 53.1 & 31.4 & 89.1 & 69.7 & \underline{88.2} & \textbf{67.7} & \textbf{0.0} & \underline{40.6} & 50.8 & 83.5 & 57.4 \\
        Ours (Real 3D) & \underline{61.7} & \underline{90.2} & \underline{48.2} & \underline{84.3} & \underline{89.4} & \underline{49.5} & 53.5 & 56.8 & \underline{36.1} & \underline{91.5} & \underline{74.6} & 79.8 & 63.4 & \textbf{0.0} & 40.3 & \underline{72.8} & \underline{83.8} & \underline{62.9} \\
        Ours (Recon) & \textbf{63.4} & \textbf{92.7} & 41.1 & \textbf{85.9} & \textbf{89.8} & \textbf{50.6} & \underline{56.3} & \textbf{61.6} & \textbf{36.3} & \textbf{94.0} & \textbf{78.6} & \textbf{90.3} & \underline{66.4} & \textbf{0.0} & \textbf{41.0} & 72.6 & \textbf{85.8} & \textbf{64.7} \\
        \midrule
        \multicolumn{1}{@{}l}{\emph{Scribble annotations \& EoMT backbone}} \\
        EMA & 62.58 & - & 47.23 & 83.63 & 89.17 & 47.51 & \textbf{62.86} & 59.52 & \textbf{36.29} & 91.19 & 71.78 & 86.77 & 61.16 & 0.00 & 42.42 & \textbf{78.58} & 82.95 & 60.31 \\
        Ours (Recon) & \textbf{63.94} & - & \textbf{47.29} & \textbf{85.76} & \textbf{89.31} & \textbf{49.27} & 57.66 & \textbf{61.79} & 34.49 & \textbf{94.05} & \textbf{78.94} & \textbf{88.25} & \textbf{66.41} & 0.00 & \textbf{43.07} & 76.62 & \textbf{85.65} & \textbf{64.47} \\
        \bottomrule
    \end{tabular}}
\end{table}

\begin{table*}[h]
    \hspace{1cm}
    \centering
    \caption[Comparison of mIoU Results and Relative Scores on Cityscapes Dataset]{\textbf{Comparison of mIoU Results and Relative Scores on Cityscapes Dataset} Bold numbers indicate the best and underlined values the second-best performance among non-fully supervised approaches.}
    \label{tab:cityscapes_comparison}
    \resizebox{\textwidth}{!}{\begin{tabular}{l|cc|*{19}{c}}
        \toprule
        \textbf{Method} & \textbf{mIoU} & \textbf{SS/FS (\%)} & \rotatebox{90}{\textbf{bicycle}} & \rotatebox{90}{\textbf{building}} & \rotatebox{90}{\textbf{bus}} & \rotatebox{90}{\textbf{car}} & \rotatebox{90}{\textbf{fence}} & \rotatebox{90}{\textbf{motorcycle}} & \rotatebox{90}{\textbf{person}} & \rotatebox{90}{\textbf{pole}} & \rotatebox{90}{\textbf{rider}} & \rotatebox{90}{\textbf{road}} & \rotatebox{90}{\textbf{sidewalk}} & \rotatebox{90}{\textbf{sky}} & \rotatebox{90}{\textbf{terrain}} & \rotatebox{90}{\textbf{traffic light}} & \rotatebox{90}{\textbf{traffic sign}} & \rotatebox{90}{\textbf{train}} & \rotatebox{90}{\textbf{truck}} & \rotatebox{90}{\textbf{vegetation}} & \rotatebox{90}{\textbf{wall}} \\
        \midrule
        Fully Supervised & 77.6 & - & 74.5 & 92.4 & 86.4 & 94.5 & 60.1 & 62.7 & 78.9 & 58.9 & 60.4 & 98.2 & 85.4 & 94.8 & 65.7 & 65.5 & 74.8 & 80.4 & 82.0 & 92.3 & 66.3 \\
        \midrule
        \multicolumn{1}{@{}l}{\emph{Point annotations}} \\
        EMA & 50.5 & 65.1 & 49.0 & 74.6 & \underline{57.8} & \underline{82.9} & 29.1 & 25.0 & 51.8 & 25.9 & 30.3 & \underline{94.1} & \underline{64.9} & 78.4 & 37.7 & 16.9 & 30.0 & \underline{52.3} & 49.4 & 77.2 & 32.7 \\
        SASFormer & 42.7 & 55.0 & 35.5 & 67.5 & 53.1 & 70.1 & \underline{33.8} & 17.8 & 38.3 & 12.1 & 10.4 & 91.8 & 41.1 & 75.8 & 18.6 & 10.8 & 25.4 & 45.8 & \underline{54.2} & 69.4 & \underline{41.5} \\
        TEL & \underline{53.0} & \underline{68.3} & \textbf{59.0} & \underline{81.2} & 39.7 & 80.6 & \textbf{36.7} & \textbf{31.3} & \textbf{62.9} & \textbf{32.8} & \textbf{44.2} & 93.2 & 62.7 & \underline{80.5} & \underline{41.7} & \textbf{36.5} & \textbf{57.8} & 30.9 & 23.9 & \textbf{84.1} & 26.8 \\
        Ours (Recon) & \textbf{56.5} & \textbf{72.8} & \underline{54.0} & \textbf{82.9} & \textbf{69.6} & \textbf{85.3} & 33.1 & \underline{31.1} & \underline{55.3} & \underline{32.4} & \underline{31.8} & \textbf{94.6} & \textbf{67.1} & \textbf{84.2} & \textbf{43.1} & \underline{25.5} & \underline{40.6} & \textbf{62.1} & \textbf{55.3} & \underline{81.6} & \textbf{43.1} \\
        \midrule
        \multicolumn{1}{@{}l}{\emph{Scribble annotations}} \\
        EMA & 61.2 & 78.9 & 63.6 & 82.2 & \underline{73.9} & \underline{83.4} & 41.7 & \underline{48.6} & 64.5 & 40.0 & 50.9 & 87.1 & 50.6 & 80.2 & 45.3 & 40.4 & 51.4 & \underline{61.3} & \underline{68.7} & 80.7 & 48.3 \\
        SASFormer & 55.6 & 71.7 & 53.9 & 82.1 & 68.4 & 74.8 & 41.4 & 23.4 & 52.7 & 26.5 & 28.5 & \underline{92.3} & 47.0 & 88.0 & 49.9 & 35.6 & 49.2 & 56.2 & 55.8 & 84.4 & 46.4 \\
        TEL & \underline{64.4} & \underline{83.0} & \textbf{69.6} & \underline{86.7} & 62.6 & 65.0 & \underline{46.7} & 47.0 & \textbf{71.2} & \textbf{48.7} & \underline{51.9} & 91.3 & \underline{58.6} & \textbf{92.5} & \textbf{53.0} & \textbf{59.1} & \textbf{68.7} & 60.1 & 54.2 & \textbf{87.6} & \underline{48.8} \\
        Ours (Recon) & \textbf{68.1} & \textbf{87.8} & \underline{65.4} & \textbf{87.5} & \textbf{78.9} & \textbf{89.0} & \textbf{48.2} & \textbf{49.6} & \underline{69.8} & \underline{47.4} & \textbf{55.2} & \textbf{94.6} & \textbf{66.9} & \underline{90.6} & \underline{52.0} & \underline{51.4} & \underline{61.6} & \textbf{68.7} & \textbf{75.8} & \underline{86.4} & \textbf{55.0} \\
        \midrule
        \multicolumn{1}{@{}l}{\emph{Coarse annotations}} \\
        EMA & \underline{66.5} & \underline{85.7} & 63.6 & \underline{87.7} & \underline{73.6} & \underline{87.8} & 49.8 & 50.7 & 64.6 & \underline{43.1} & 47.0 & \textbf{95.9} & \textbf{71.7} & \underline{91.9} & \underline{52.6} & 50.2 & 60.1 & \underline{59.8} & \underline{72.6} & \underline{87.5} & \underline{53.0} \\
        SASFormer & 42.8 & 55.2 & 41.1 & 68.6 & 54.5 & 74.3 & 33.8 & 15.0 & 40.8 & 7.8 & 16.4 & 92.7 & 37.4 & 79.6 & 14.2 & 9.1 & 26.0 & 36.4 & 56.2 & 70.2 & 39.2 \\
        TEL & 64.9 & 83.7 & \textbf{69.9} & 87.3 & 62.8 & 73.1 & \underline{50.9} & \textbf{56.3} & \textbf{70.4} & \textbf{49.6} & \textbf{54.2} & 91.3 & 55.8 & 91.1 & 49.5 & \textbf{61.9} & \textbf{71.2} & 59.5 & 51.0 & \underline{87.5} & 39.7 \\
        Ours (Recon) & \textbf{68.6} & \textbf{88.4} & \underline{64.7} & \textbf{89.0} & \textbf{80.6} & \textbf{88.3} & \textbf{51.2} & \underline{53.1} & \underline{67.3} & \underline{43.1} & \underline{48.8} & \underline{95.7} & \underline{69.9} & \textbf{92.5} & \textbf{55.7} & \underline{52.3} & \underline{62.6} & \textbf{67.7} & \textbf{77.9} & \textbf{87.7} & \textbf{55.2} \\
        \midrule
        \multicolumn{1}{@{}l}{\emph{Scribble annotations \& EoMT backbone}} \\
        EMA & 71.15 & - & 64.96 & 86.88 & 87.56 & 89.79 & 52.10 & 54.00 & \textbf{72.72} & 50.65 & 57.58 & \textbf{95.71} & \textbf{73.52} & 88.77 & 53.06 & 53.93 & 62.45 & 79.96 & 85.03 & 85.49 & 57.67 \\
        Ours (Recon) & \textbf{73.50} & - & \textbf{69.81} & \textbf{88.86} & \textbf{88.81} & \textbf{91.00} & \textbf{58.77} & \textbf{62.64} & 72.18 & \textbf{52.17} & \textbf{58.59} & 95.59 & 71.86 & \textbf{90.80} & \textbf{56.04} & \textbf{55.68} & \textbf{66.45} & \textbf{82.45} & \textbf{87.17} & \textbf{86.66} & \textbf{60.93} \\
        \bottomrule
    \end{tabular}}
\end{table*}

\clearpage
\section{Sampling Strategies for Reconstruction}\label{supp:CMCsampling}
\begin{figure*}[h]
    \centering
    \vspace{1cm}
    \includegraphics[width=\linewidth]{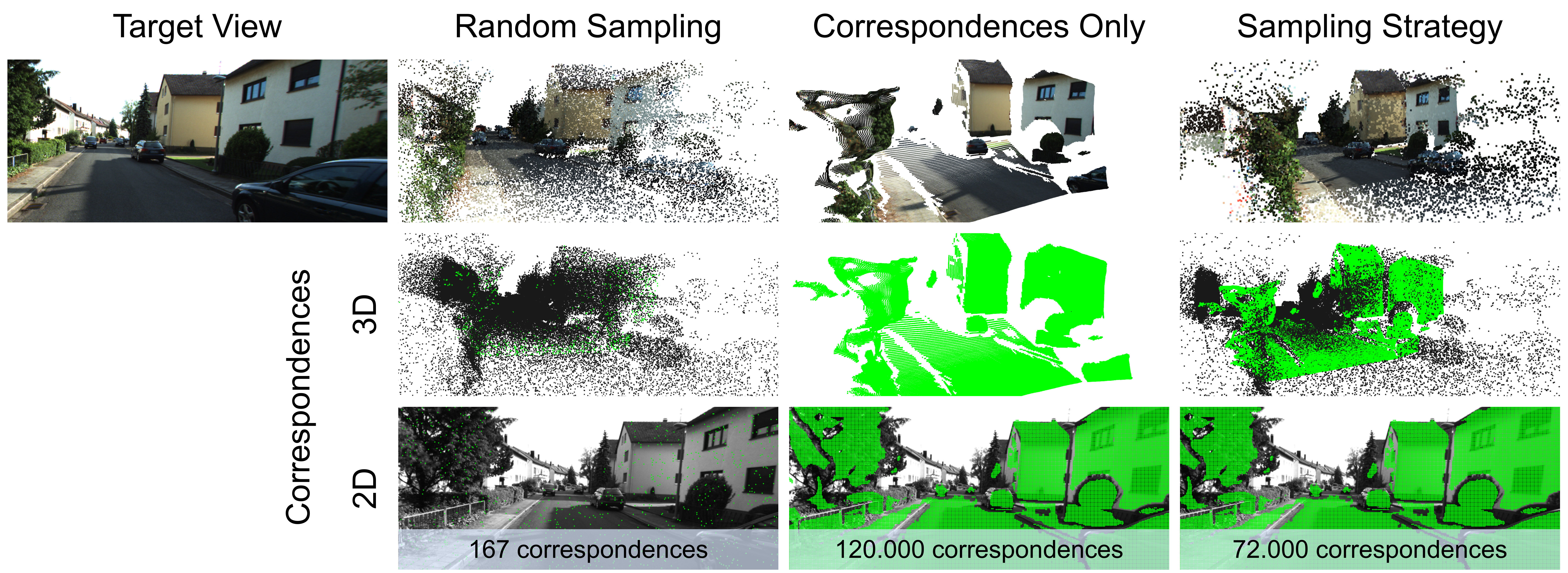}
    \caption{\textbf{Visualization of sampling strategies for cross-modal consistency.} 
    The rows display the 3D point cloud (middle) and its projection onto the 2D target view (bottom). Green points indicate valid 2D-3D correspondences for the target image, while black points represent the remaining scene geometry.
    Left: Random Sampling (120k points global) results in sparse correspondences (avg. 167) in the view, insufficient for dense consistency learning. 
    Center: Sampling strictly from correspondences (100\% target view) maximizes alignment but results in a fragmented 3D scene (note the missing car geometry), preventing the 3D network from learning global context. 
    Right: Our View-Aware Sampling Strategy (60\% target view, 40\% spatial context) ensures dense correspondences ($\sim$72K) while maintaining the holistic scene structure required for robust 3D segmentation.}
    \label{fig:supp_sampling_strategy}
\end{figure*}

As discussed in the main paper, efficiently processing the massive point clouds generated from video sequences (often exceeding 60 million points) presents a challenge. Simply downsampling the global point cloud randomly is ineffective for our Cross-Modal Consistency (CMC) loss, as it yields too few correspondences (approx. 167 points) within the target image's field of view, as shown in Fig.~\ref{fig:supp_sampling_strategy}. Conversely, retaining points exclusively from the target view maximizes correspondences but discards the geometric context necessary for the 3D network to learn robust features, resulting in fragmented 3D shapes (see ``Correspondences Only'' in Fig.~\ref{fig:supp_sampling_strategy}). To resolve this, we employ a \textit{View-Aware Sampling Strategy} that constructs a hybrid point cloud: 60\% of points are sampled from the current camera view to ensure dense 2D-3D alignment for supervision, while the remaining 40\% are sampled from the surrounding scene to provide structural context. This balanced approach enables both effective cross-modal transfer and accurate 3D segmentation.

\clearpage
\section{Analysis of Real vs. Reconstructed Supervision}\label{supp:real-vs-recon}

In \cref{sec:main_results} of the main paper, we observed the counterintuitive result that supervision from reconstructed point clouds ("Ours (Recon)") outperformed ground-truth LiDAR ("Ours (Real 3D)") on the Waymo dataset (53.3\% vs. 51.8\%). We hypothesized that this performance gap stems from two factors inherent to the raw LiDAR data: (1) high sparsity leading to fewer valid correspondences, and (2) the lack of a reconstruction confidence score, which forces the use of single-term weighting rather than our proposed Dual Confidence mechanism.

To validate this hypothesis, we conducted a controlled experiment where we artificially degraded our Reconstructed data to mimic the characteristics of the Real LiDAR sensor. Specifically, we:
\begin{enumerate}
    \item Restricted the reconstruction to single-scan density
    \item Reduced the amount of correspondences to the average amount of correspondences available in the real 3D data
    \item Disabled the reconstruction confidence weighting component, relying only on prediction confidence (matching the Real 3D setup).
\end{enumerate}

The results are presented in Table~\ref{tab:lidar_simulation}. When the reconstructed data is restricted to the same sparsity and weighting constraints as the Real LiDAR ("Recon (Simulated LiDAR)"), the performance drops to 51.7\%, virtually matching the Real 3D performance of 51.8\%. This confirms that the superior performance of \method stems specifically from the density of the multi-view reconstruction and the noise-suppression capability of the reconstruction confidence, rather than an artifact of the data source itself.

\begin{table}[h]
    \centering
    \vspace{1cm}
    \caption{\textbf{Validation of Real vs. Reconstructed Gap (Waymo).} We simulate the characteristics of Real LiDAR (sparsity and lack of confidence scores) using our Reconstructed data. When matching the constraints of Real LiDAR, our performance aligns with the Real 3D baseline, confirming that the gains of \method come from the increased density and dual-confidence weighting enabled by the reconstruction.}
    \label{tab:lidar_simulation}
    \footnotesize
    \begin{tabular}{lcccc}
        \toprule
        \textbf{Source} & \textbf{Density} & \textbf{Confidence Weighting} & \textbf{mIoU (\%)} \\
        \midrule
        Ours (Real 3D) & Sparse (LiDAR) & Single (Pred. Only) & 51.8 \\
        \midrule
        Ours (Recon) & Sparse (Simulated) & Single (Pred. Only) & 51.7 \\
        \textbf{Ours (Recon)} & \textbf{Dense (Multi-view)} & \textbf{Dual (Pred. + Rec.)} & \textbf{53.3} \\
        \bottomrule
    \end{tabular}
\end{table}
\clearpage
\section{Label Generation}\label{supp:data-gen}
The scribble labels employed in the majority of our evaluations were generated using Scribbles~for~All~\cite{boettcher_scribbles_2024}. While we utilized the published labels for KITTI-360 and Cityscapes, the annotations for Waymo and NYUv2 were generated specifically for this work. Table~\ref{tab:scribble_configs} details the configuration parameters adopted for the label generation process.

\begin{table*}[h]
\centering
\vspace{1cm}
\caption[Scribble Generation Configurations for Waymo Open Dataset and NYUv2]{\textbf{Scribble Generation Configurations for Waymo Open Dataset and NYUv2}}
\footnotesize
\begin{tabular}{l|c|c}
\toprule
\textbf{Parameter} & \textbf{Waymo Open Dataset Value} & \textbf{NYUv2 Dataset Value} \\
\midrule
height distortion & 0.9 & 1 \\
min binary erosion & 2 & 2 \\
max binary erosion & 40 & 20 \\
it extra erosion & 5 & 2 \\
min erosion area share & 0.002 & 0.003 \\
max erosion area share & 0.05 & 0.15 \\
background px value & [0, 0, 0] & [255, 255, 255] \\
background input values & [0] & [0] \\
ignore values & [0] & [255] \\
patience & 20 & 20 \\
error tolerance px & 0 & 0 \\
min blob area & 1500 & 80 \\
line thickness & 5 & 3 \\
scribble scale & 1.0 & 1.0 \\
\bottomrule
\end{tabular}
\label{tab:scribble_configs}
\end{table*}
\clearpage
\section{Implementation Details \& Hyperparameters}\label{supp:implementation}
This section provides a comprehensive overview of the hyperparameters and augmentation strategies employed for the 2D and 3D branches of our framework. Table~\ref{tab:hyperparameters} specifies the optimization settings, which are aligned with standard practices for SegFormer and Point Transformer V3 to ensure fair comparisons. Additionally, Tables~\ref{tab:aug_2d} and~\ref{tab:aug_3d} itemize the specific data augmentations applied. Notably, we apply stronger augmentations to the student models relative to the teachers—a fundamental aspect of the Mean Teacher paradigm designed to enforce consistency and robustness.

\begin{table}[h]
    \centering
    \caption{\textbf{Overview of core training hyperparameters for 2D and 3D branches}}
    \label{tab:hyperparameters}
    \footnotesize
    \begin{tabular}{ll|ll}
        \toprule
        \multicolumn{2}{c}{\textbf{2D Branch}} & \multicolumn{2}{c}{\textbf{3D Branch}} \\
        \textbf{Config} & \textbf{Value} & \textbf{Config} & \textbf{Value} \\
        \midrule
        Optimizer & AdamW & Optimizer & AdamW \\
        Scheduler & PolyLR & Scheduler & OneCycleLR \\
        Criteria & CrossEntropy & Criteria & CrossEntropy \\
        ~ & Student Teacher Loss & ~ & Student Teacher Loss \\
        Learning Rate & 0.00005 & Learning Rate & 0.001 \\
        Weight Decay & 1e-08 & Weight Decay & 0.005 \\
        Batch Size & 12 & Batch Size & 12 \\
        Datasets & KITTI-360 / & Datasets & KITTI-360 / \\
        ~ & Waymo / & ~ & Waymo / \\
        ~ & NYUv2 / & ~ & NYUv2 / \\
        ~ & Cityscapes & ~ & Cityscapes \\
        Epochs & 50 (250 NYUv2) & Epochs & 50 (250 NYUv2) \\
        \bottomrule
    \end{tabular}
\end{table}

\begin{table}[h]
    \centering
    \caption{\textbf{Overview of Applied Data Augmentations in 2D.} The first two columns list the augmentations and their corresponding parameters. The third and fourth columns indicate whether each augmentation is applied to the inputs of both the student and teacher models, or to only one of them.}
    \label{tab:aug_2d}
    \footnotesize
    \begin{tabular}{llcc}
        \toprule
        \textbf{Augmentation} & \textbf{Parameters} & \textbf{Student} & \textbf{Teacher} \\
        \midrule
        Random Horizontal Flip & p: 0.5 & \checkmark & \checkmark \\
        Scale and Distort & scale: [0.5, 1.2], distort: [0.9, 1.1] & \checkmark & \checkmark \\
        Random Crop & KITTI-360: (376, 512) & \checkmark & \checkmark \\
        ~ & Waymo: (640, 640) & ~ & ~ \\
        ~ & NYUv2: (480, 480) & ~ & ~ \\
        Gaussian Blur & kernel size: 7; p: 0.5 & \checkmark & \\
        Augmix & severity: 2; p: 0.5 & \checkmark & \\
        Cutout (1) & square size: 90; p: 1.0 & \checkmark & \\
        Cutout (2) & square size: 90; p: 0.5 & \checkmark & \\
        \bottomrule
    \end{tabular}
\end{table}

\begin{table}[h]
    \centering
    \caption{\textbf{Overview of Applied Data Augmentations in 3D.} The first two columns list the augmentations and their corresponding parameters. The third and fourth columns indicate whether each augmentation is applied to the inputs of both the student and teacher models, or to only one of them.}
    \label{tab:aug_3d}
    \footnotesize
    \begin{tabular}{llcc}
        \toprule
        \textbf{Augmentation} & \textbf{Parameters} & \textbf{Student} & \textbf{Teacher} \\
        \midrule
        Random Rotation & axis: z, angle [-1, 1]; p: 0.5 & \checkmark & \checkmark \\
        Random Scale & scale: [0.9, 1.1] & \checkmark & \checkmark \\
        Random Flip & p: 0.5 & \checkmark & \checkmark \\
        Random Jitter & sigma: 0.005; clip: 0.02 & \checkmark & \\
        PointClip & [100, 40, 100] & \checkmark & \\
        \bottomrule
    \end{tabular}
\end{table}
\clearpage
\section{Colormaps}\label{supp:colormap}
\begin{figure*}[h]
    \centering
    \includegraphics[width=0.8\linewidth]{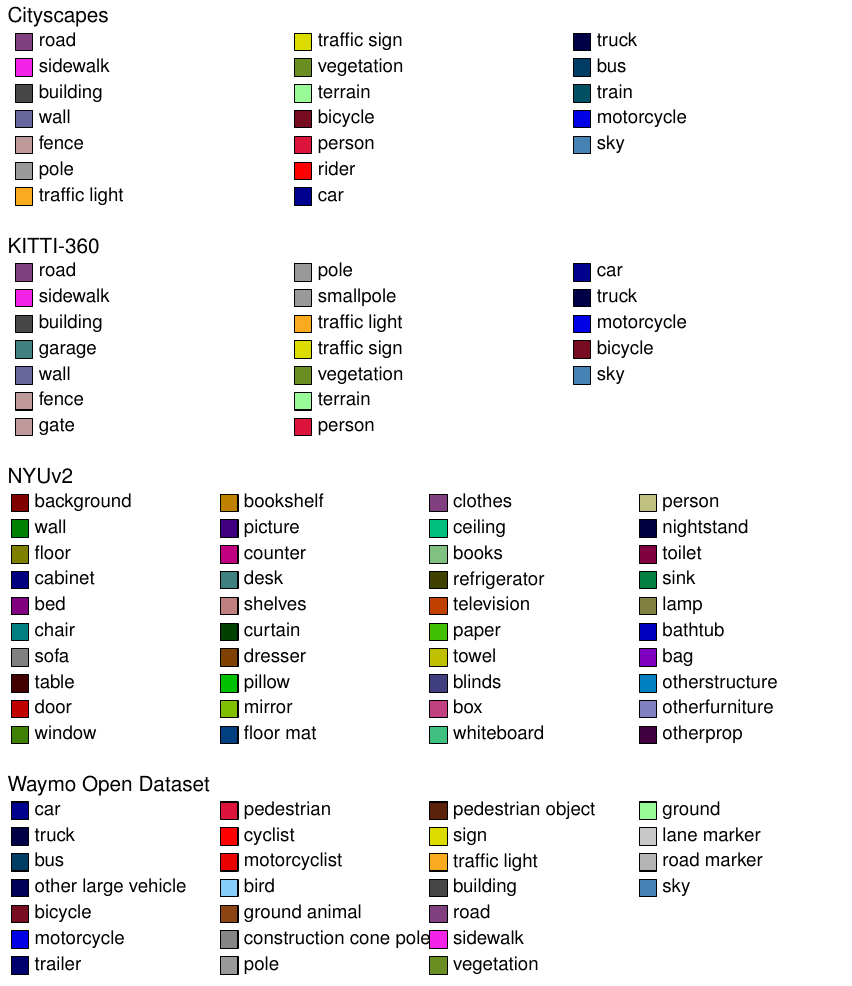}
    \caption{\textbf{Semantic class color coding for all evaluated datasets.} We visualize results using consistent color mappings across each dataset: Cityscapes (19 classes), KITTI-360 (19 classes), NYUv2 (40 classes), and Waymo Open Dataset 
    (25 classes). Color assignments follow the official dataset conventions where 
    available. Note that while some classes share similar names across datasets 
    (e.g., 'road', 'car', 'person'), their definitions and label protocols may 
    differ between datasets.}
\end{figure*}
\clearpage

\end{document}